\documentclass[manuscript,screen]{acmart}

\usepackage{graphicx}
\usepackage{lineno}
\usepackage{graphicx}
\usepackage{amsmath}
\usepackage{longtable}
\usepackage{array}
\usepackage{pdflscape}
\usepackage{ctable}
\newcolumntype{L}{>{\centering\arraybackslash}m{7cm}}
\newcolumntype{S}{>{\centering\arraybackslash}m{4cm}}
\newcolumntype{M}{>{\centering\arraybackslash}m{6cm}}
\DeclareMathOperator*{\argmax}{argmax}

\AtBeginDocument{%
  \providecommand\BibTeX{{%
    \normalfont B\kern-0.5em{\scshape i\kern-0.25em b}\kern-0.8em\TeX}}}

\acmDOI{10.1145/3527448}

\begin{document}

\title{Explainable Deep Reinforcement Learning: State of the Art and Challenges}

\author{George A. Vouros}
\email{georgev@unipi.gr}
%\email{webmaster@marysville-ohio.com}
\affiliation{
  \institution{University of Piraeus, Greece}
  \streetaddress{Gr Lambraki 126}
  \city{Piraeus}
  \state{Greece}
  \postcode{18534}
}

\renewcommand{\shortauthors}{George A. Vouros}

\begin{abstract}
  Interpretability, explainability and transparency are key issues to introducing Artificial Intelligence methods in many critical domains: This is important due to ethical concerns and trust issues strongly connected to reliability, robustness, auditability and fairness, and has important consequences towards keeping the human in the loop in high levels of automation, especially in critical cases for decision making, where both (human and the machine) play important roles. While the research community has given much attention to  explainability of closed (or black) prediction boxes, there are tremendous needs for explainability of closed-box methods that support agents to act autonomously in the real world. Reinforcement learning methods, and especially their deep versions, are such closed-box methods. In this article we aim to provide a   review of state of the art methods for explainable deep reinforcement learning methods, taking also into account the needs of human operators - i.e., of those that take the actual and critical decisions in solving real-world problems. We provide a formal specification of the deep reinforcement learning explainability problems, and we identify the necessary components of a general explainable reinforcement learning framework. Based on these, we provide a comprehensive review of state of the art methods, categorizing them in classes according to the paradigm they follow, the interpretable models they use, and the surface representation of explanations provided. The article concludes identifying open questions and important challenges.
\end{abstract}

\begin{CCSXML}
<ccs2012>
   <concept>
       <concept_id>10010147.10010257.10010258.10010261</concept_id>
       <concept_desc>Computing methodologies~Reinforcement learning</concept_desc>
       <concept_significance>500</concept_significance>
       </concept>
 </ccs2012>
\end{CCSXML}

\ccsdesc[500]{Computing methodologies~Reinforcement learning}

\keywords{Deep Learning, Deep Reinforcement Learning, Interpretability, Explainability, Transparency}

\maketitle

%\linenumbers

\section{Introduction}
\label{intro}

EU ethical guidelines \cite{EUEthicalGuidelines} on accountability, technical robustness and safety,
oversight, privacy and data governance, non-discrimination and fairness, transparency, societal and environmental
well-being specify the essential aspects to be considered by Artificial Intelligence (AI) and Machine Learning (ML) products and appliances. Concise analysis for accredited  adherence of systems to EU ethical principles is an essential prerequisite for the acceptability of deploying and using any such application. In addition, the U.S.A National Security Commission reports in \cite{AiCommission} that  ``\textit{Ethical considerations are an important facet of R\&D, application, training, protection, and cooperation in AI. ... Developing trustworthy AI systems is essential for operational integrity and adoption. It is closely connected to, and depends on, reliability, robustness, auditability, explainability and fairness. From the earliest phase, systems should be designed with ethics in mind}".

Designing and implementing systems with ``deep knowledge" prevails in AI: We may recall the need of incorporating ``deep knowledge" about the domains and problems addressed (in contrast to ``shallow knowledge'') in  expert and knowledge-based systems, mitigating brittleness, vulnerability to unforeseen circumstances,  lack of ability to reason from ``first principles" when needed,  and inability to provide informative explanations on the reasoning towards reaching conclusions and prescribing solutions.  
Deep knowledge is indeed essential for AI systems, nowadays  with some very emergent, important and tremendously larger/broader challenges, and with more severe, challenging consequences and inescapable needs: In contrast to interpretable systems that exploit explicit knowledge and whose reasoning mechanisms  principles and  inner-workings are more transparent to humans, today deep knowledge is encoded in hidden layers of multi-layered deep neural networks, in the form of weights on ``synapses", or in models that fit a value function ranging into sets  of large number of actions in millions of states. Reaching this point of technological advances allows us to deploy autonomous systems acting in the real-world with significantly more advanced levels of perception, decision making and automation, all integrated in a single loop. These systems have not been instructed by direct advice, but  have been trained on massive data sampled  from the real world. Most importantly, the interpretability of knowledge and reasoning mechanisms has been lost, not only for the domain experts, or for the humans using these systems, but also for AI systems' engineers and developers. This has huge impact on ethical concerns when deploying and using these systems, thus  on systems' accountability, liability, safety and trustworthiness.%\footnote{Not to mention the irrational exercise where humans have to develop trust on algorithms that ``invent" models about the world without being able to explain what they have learned and why. }. 

The consequences are crucial especially when we aim at higher levels of automation, where models aim to support autonomous agents to act autonomously.  The problem is addressed in the literature from different perspectives, depending on the targeted use of explanations: Machine learning experts and system developers need explanations of algorithms' inner workings, while domain experts and operators using/working with autonomous systems need to understand how the system reaches specific decisions, especially when the solutions proposed are not intuitive or obvious, understanding models' ``whims" (to use an anthropocentric metaphor), objectives and capabilities.  

Towards developing trustworthy and accountable autonomous systems (agents) to support automation in safety-critical real-world settings, this article aims to review state of the art methods regarding   interpretability, explainability and transparency of reinforcement learning (RL) methods, with emphasis on their deep (DRL) versions: Currently we lack a concise understanding of the qualities of explainable and transparent DRL methods and we have to close this gap. This is important  in many real-world complex settings in which DRL agents are deployed, spanning from (co-)driving autonomous vehicles, to supporting automation in air traffic management and other cases with time-critical decisions.
The challenges to make  DRL models interpretable/explainable, are great: Some of these are ``inherited" from ML models in general, but for DRL methods some additional subtle and important issues need to be considered, given that DRL models exploit intertwined closed prediction box models and they do solve sequential decision tasks, aiming to increase the expected feature reward w.r.t. agents' preferences, intended objectives, and constraints that the environment imposes (e.g. partial observability, delayed rewards, dynamics, multiple co-existing agents, etc):

$\bullet$ Agents are expected to provide informative explanations even in cases where humans cannot explain their own way of thinking / acting on solving sequential decision problems, comprehensively and with sufficient clarity: This is for instance true in cases where it is easier for humans to demonstrate how they apply their knowledge and skills in example cases. For instance, when systems follow a data-driven approach learning from demonstrations, it may be hard to explicate the acquired knowledge. In these cases, providing concise ``mainstream" explanations (e.g., explicating their knowledge and reasoning chains for expected rewards in long-term time horizons) may simply fail: Systems must be empowered with sophisticated means that allow them to demonstrate reasoning and problem solving abilities more intelligently (i.e., taking into account informative factors, providing examples of behaviour, or critical aspects of decision making), supporting users to understand systems' rationale at various levels of detail. Let me provide an example on this. We may all have used the "shoe laces" example in our early AI lectures to show how difficult it is for experts to express procedural knowledge on specific sequential tasks and problem solving, and thus, how difficult it is for systems to represent in symbolic and explicit ways such knowledge. However, nowadays, providing paradigmatic demonstrations of expert skills and activity may be enough for a system to gain competency in performing such tasks and in solving sequential decision problems. Such a system should explicate facets of the knowledge acquired and must have the ability to provide comprehensive explanation for how solutions are reached along trajectories of states reached, how objectives are achieved, and when/why it fails. 

$\bullet$ We need to investigate what  the principles or paradigms of building an interpretable/explainable DRL system are: Is it enough to add some interpretability/explainability modules to an DRL system (as we would do in any ML system), or do we need to design a system from the early phases to be explainable (as it is conjectured to be the case for building collaborative and ethical systems)? In other words, do we need to design and develop inherently explainable DRL methods? An answer to this question may vary between domains and conditions of systems' deployment, but this decision should not be based on any assumption about a trade-off between  accuracy in prediction and interpretability. While authors in \cite{PrinciplesPracticeofXML} propose methodological aspects for reaching such a decision for explainable machine learning, authors in \cite{InterpretableML} conjecture that closed boxes are generally unnecessary, given that their accuracy is generally not better than a well designed interpretable model (whose probability of existence is rather high). However, a DRL method may comprise more than one interpretable models whose functionality is intertwined, and whose interpretability / accuracy may affect  final outcomes. 

$\bullet$ We need to answer ``what" information and ``how" it should be communicated for providing effective explanations. We may take advantage of research in other scientific fields (social and cognitive sciences, as well as in philosophy) regarding the qualities of ``good" explanations. Towards this, we need to consider not only the qualities of the explanation content (``what"), but also the modalities that will be used to provide surface representations of that information (``how"), the characteristics of the context in which explanations are provided, and the purpose that they need to serve w.r.t. constraints and requirements (pragmatic aspects). E.g., humans may be acquainted with specific modes and forms of information, using specific nomenclature and/or visual means, also in relation to responsiveness, temporal requirements in certain contexts, cognitive load, etc.

$\bullet$ We need concise experimental protocols for explainability, considering the purposes and contexts of explanations' provision together with human factors, in relation to the explanation problem and the required qualities of explanations. This will allow the community working on explainability to compare different approaches, and build a firm background, making conjectures and refuting theories on explainability, proceeding on more advanced results. While this holds for ML methods in general \cite{ExplanationsQuality}, experimental protocols are tremendously important for DRL methods, which have to support humans to infer faithful models of behaviour, objectives and capabilities of agents, towards developing trustworthiness in real-world problem solving settings.  

Motivated by these challenging issues, in this article we review explainable DRL (XDRL) methods. We aim to   understand their potential capabilities and limitations to solve interpretability/explainability problems while focusing on the needs of human operators. Operators need to be  aware not only  of the particular situation the agent faces, but also of the role that certain real-world features play in decision making, of the agent's objectives, decision making capabilities and preferences to act, of the overall policy, and of the choices made under specific (potentially, critical) circumstances.  In general, operators need to be able to understand and judge in an informed way when to trust an agent, when to take control and/or apply appropriate mitigation actions, and when to provide further advice for refining agent's knowledge. 
While surveys on XRL exist (e.g. \cite{SurveyXRL},\cite{Multidisciplinary}), here we aim at a comprehensive survey focusing on XDRL, providing details about all (according to our knowledge)  works describing and  providing well-justified insights on state of the art methods, under a common framework, distinguishing clearly between existing paradigms, providing details on evaluating these approaches, also with respect to transparency, and stating emerging and important  challenges.

Structure of this article: In Section 2, we  provide preliminary knowledge and discuss interpretability / explainability aspects that we must consider, as well as explainability desiderata. Section 2 formulates the problems of providing explanations for DRL agents. Based on the definitions provided, Section 3 identifies the functionality of the necessary components of an XDRL method and attempts a  categorization of XDRL methods, based on the explanation  paradigm they follow and needs they do address. Section 4 provides a comprehensive review of the state of the art methods, and Section 5   concludes the paper with final conclusions and challenges ahead. The Appendix (supplementary online material) of this article provides a detailed categorization of reviewed methods in three Tables, using a comprehensive set of characteristics.

\section{Interpretability, Explainability and Transparency  of DRL methods}
\label{sec:formulation}

This section sets the ground to proceed towards formulating the explainability problems for DRL agents, and providing a general framework of XDRL agents in the next section.
First, it provides DRL  preliminary knowledge, so as to introduce terminology and symbols that we will use in subsequent sections, assuming  that the readers are acquainted with DRL methods. It proceeds to  clarify the notions of  explainability, interpretability and transparency, which is useful towards understanding the various facets of the problem, and states explainability desiderata. Finally, it provides definitions of the expainability problems for DRL agents.

\subsection{DRL Preliminaries}
\label{sec:preliminaries}

Given a set of states $S$ and a set of actions $A$ available to an agent acting in an environment $E$, possibly with continuous parameters, and a reward function $r: (S \times A; E) \rightarrow \mathbb{R}$, the goal of the agent is to learn a policy $\pi$, i.e., a mapping from states to actions or to a distribution over actions, that will maximize its expected cumulative rewards. 

Policy and reward models $\pi_{\theta}$ and $r_{\phi}$ are parameterized by variables' vectors $\theta$ and $\phi$, respectively, and take as input subsets of $M$, a set of real-world features.
A trajectory in time horizon (length) $H$, is defined as a sequence of states $s_t$ and actions $a_t$, for $t=1,\cdots,H$. The policy and reward parameters determine the optimal trajectory $\tau^{\theta, \phi}_{E}$ in the environment $E$:
\[\tau^{\theta, \phi}_{E} = \argmax_{\tau^{\theta}_E \in TR^{\theta}_E}   \sum_{t=0}^H\gamma^t r_{\phi}(s_t, a_t; E) \]
\noindent where  $\gamma$ is the discount factor between 0 and 1 that controls the trade-off between immediate and future rewards, and $TR^{\theta}_E$ denotes the set of all possible trajectories in environment $E$, given parameters $\theta$ and reward function $r_{\phi}$. Those are the trajectories that can be sampled from the policy $\pi_\theta$ (i.e., $\tau^{\theta}_E \sim \pi_{\theta}$). 

The goal of DRL methods is, given the reward function $r_\phi,$ to tune the parameters $\theta$ so as to maximize the expected cumulative reward while acting in the environment $E$, i.e., determine $\theta^*$, such that
\[\theta^*=\argmax_{\theta} \mathbb{E}_{\tau^{\theta}_E \sim \pi_{\theta}} \sum_{t=0}^H\gamma^t r_{\phi}(s_t, a_t; E) \]
The agent, depending on the RL method used, may be required to fit a model of the environment $E$ and/or estimate the values of states, or the values of state-action pairs (i.e., the expected cummulative reward from that state, or state and action, respectively), or even (e.g in case of imitation learning) to learn a model of the reward function, tuning parameters $\phi$, given the set of states' features $M$.

\subsection{Explainability, interpretability and transparency}
\label{sec:interpretability}

Given the above specifications and the  models that a DRL method comprises (policy, reward models, etc - these are detailed in subsequent sections), we need to specify what  it means to enhance a DRL method with explainability and interpretability, providing transparency in decision making. 

Lipton \cite{Lipton} asserts that explanation is post-hoc interpretability. 
Biran and Cotton \cite{BiranCotton} (among others, e.g. C. Rudin at al. in \cite{InterpretableML}) define  interpretability  as the degree to which one can understand the response of a system. Relevance, of information for relationships contained in the data or learned by a machine learning model, for a particular audience into a chosen problem,  is an important aspect of interpretable models in \cite{Murdoch22071}.  Explanation is a particular way of obtaining such an understanding, selecting ``what" must be communicated and ``how" this is presented (e.g., using textual and/or visual means, providing ``global" vs ``local" or case-specific explanations, at different levels of detail, scale and granularity, etc.): This usually involves filtering interpretations and realizing  surface representations of interpretations, using an explanation logic that exploits interpretable models. A distinction between ``what" and ``how" is also considered in \cite{SurveyXRL}, 

To avoid overlaps and  circularity and provide a concise definition of terms, close to the definitions from \cite{SurveyXRL}, we use the term ``$explanation$" to denote the surface representation of an interpretation provided to the user. The term ``$interpretation$" is used to denote the explanation content provided by an interpretable model. Therefore,   ``$interpretability$" is the ability of a system to provide explanations' content exploiting an interpretable model, ``$explainability$" is the ability to provide surface representations of interpretations. ``$Transparency$'' is the ability of the system to generate explanations that are understandable in the context of system deployment, w.r.t. domain-specific set of constraints: This definition of transparency is close to the first principle (definition) specified in \cite{InterpretableML}. 

More specifically, considering that \emph{explainability} is the ability of systems to provide qualitative understanding of responses and objectives, given environment features from the set  $M$ and information provided by interpretable models, and close to our focus on DRL, we need to elaborate and clarify various aspects: 

$\bullet$ First, what does ``\textit{qualitative understanding}" mean; why do we require qualitative  understanding; what should be made understandable, and how can this be achieved?

$\bullet$ Then, what is a \textit{system's response}  in the context of DRL methods? Do we consider any single such response at any time instance, or sets of responses (maybe aggregated or summarized) towards the evolution of a (sub-)trajectory, reaching certain objectives? 
 
$\bullet$  What \textit{features} from the set of available features in $M$ should be considered for providing an explanation and how do we determine the importance of features?
  
$\bullet$  Finally, a critical dimension that should be considered while elaborating on any of the above questions is \textit{whose needs} an XDRL method must satisfy?

First,  qualitative understanding expresses how humans understand the qualities of an agent's policy, agent's preferences, abilities and objectives, being able to provide accurately the rationale for agent's chosen courses of actions, as the environment evolves within a state space, or for specific responses under certain circumstances. Such an understanding must occur in a human interpretable domain including domain specific concepts, entities and environment features that are presented using domain terms and other information modalities that humans are acquainted with while operating in a domain (e.g., sounds, or special visualization symbols). Such an understanding is in contrast to providing, for instance, detailed calculations, or detailed numerical assessments of any kind, e.g, by means of visualizations, at any scale and level of detail: This may of course be useful for  understanding the inner workings of any method and model quantitatively, but it cannot serve the purpose of explaining the qualities of an agent's decision making in an explicit way.
This requirement for qualitative understanding is more relevant to domain operators who need to understand why an agent takes, or why it does not take a decision in a specific circumstance, at any level of detail, granularity or scale (we explain these terms  further in the next paragraphs). Thus, such an understanding supports humans to generalize beyond individual explanations,  allowing them to infer how probable it is for the agent to  follow specific modes of behaviour, apply specific skills\footnote{Skills comprise options, macro-actions, behaviour modalities: Skills drive, in a hierarchical way, agent decisions at a low level of detail.} or take a certain action under specific circumstances. 

This is equivalent to supporting humans to infer models on (a) how the agent responds in certain circumstances w.r.t its  abilities and preferences, (b) the agent's behaviour w.r.t. its objectives, (c) the agent's behaviour w.r.t its abilities, preferences \textit{and} objectives, without considering ``where do the numbers come from". 
These models can be denoted respectively by $P(\sigma^{\theta, \phi}_E|f(\theta); E)$, $P(\sigma^{\theta, \phi}_E|f(\phi); E)$ or $P(\sigma^{\theta, \phi}_E|f(\theta), f(\phi); E)$, where $\sigma$ denotes the agent's response to the environment (e.g. an action, a (sub-)trajectory, or any skill), revealing agent's behavioural choices at different scale and granularity levels, and $f(\cdot)$ provides a representation of those model elements that drive agent's response at any time point or time period, at a low level. In other words, $f(\cdot)$ can be considered as a ``window" to the DRL agent's inner models' aspects. Such functions may, for example, provide $Q$ values, importance/saliency of state features assessed by the models, policy rollouts, even MDPs components or models' raw weights. To keep the presentation simple, we consider a single function, whose indicated arguments $\phi$ or $\theta$ indicate which of the DRL models' elements $f$ exploits.

At this point we can elaborate further on the different roles these models play: 

$\bullet$ $P(\sigma^{\theta, \phi}_E|f(\theta); E)$ models how humans interpret the agent's activity under various circumstances in $E$, based on their understanding of agent abilities and preferences. E.g., one may infer an agent's preferable mode of behavior in $E$, considering a subset of parameters $M$.

$\bullet$ $P(\sigma^{\theta, \phi}_E|f(\phi); E)$ models how humans interpret the agent's activity in the environment, given their beliefs on the objectives targeted in $E$. This allows humans to infer how the agent's objectives affect its responses. E.g., one may infer the agent's mode of behaviour given various weighted mixtures of features in $M$, according to importance.

$\bullet$ Finally, $P(\sigma^{\theta, \phi}_E|f(\theta), f(\phi); E)$ provides a model of how humans interpret agent's activity based on their beliefs on agent abilities, preferences and objectives targeted in $E$. E.g., one may infer an agent's mode of behaviour given different combinations of capabilities, preferences and environment features' importance.

Just to give a further example on these models and on their differences, given an apprenticeship learning paradigm, $P(\sigma^{\theta, \phi}_E|f(\theta); E)$   depends mainly on the training examples demonstrated to the agent;  $P(\sigma^{\theta, \phi}_E|f(\phi); E)$  depends mainly on the  set of features selected to train agent's reward model, while $P(\sigma^{\theta, \phi}_E|f(\theta), f(\phi); E)$  depends on both.

The second question concerns the granularity and scale at which the agent's responses are explained. While $granularity$  refers to the level of detail that the agent behaviour is considered, $scale$ refers to the extent where a specific response is considered. Thus granularity may refer to the actions performed at any time instance, or -in a more coarse level- to the trajectories chosen and their aggregated features, or - in even more coarse terms- to the mode of behaviour or skill exercised by the agent within a time period. Scale refers to the temporal or to the spatial (in whatever space dimensions) extent in which responses are considered. For instance, one may consider a trajectory within a specific time horizon as a response. Sub-trajectories in more restricted time horizons may be considered, while one may consider trajectories in different spatial extents and dimensions.
Thus, depending on the kind of response, $\sigma^{\theta, \phi}_E$ may be instantiated to either:
(a) $\alpha^{\theta, \phi}_E$, denoting specific actions, 
(b) $\tau^{\theta, \phi}_E$, denoting specific (sub-)trajectories, or 
(c) $\zeta^{\theta, \phi}_E$, denoting specific skills, given a set of skills $Z$, where $\zeta$ belongs.
For instance, $P(\zeta^{\theta, \phi}_E|f(\theta), f(\phi); E)$ models how humans understand an agent's choices of skills exercised in $E$, based on their understanding of agent's preferences and objectives, using the elements provided by $f(\theta), f(\phi)$. 
Finally, it must be noted that granularity and scale parameters are not included in the denotation, so as to simplify it. For instance, an action is a response given at a specific time instant $t$ and it is denoted by $\alpha_{E,t}^{\theta, \phi}$, while a trajectory in a time horizon $H$, from $t_0$ (now)  to $t_H$, is denoted by $\tau_{E,H}^{\theta, \phi}$, etc.

The third question refers to the environment features in $M$ that are being used for providing an explanation of agent's responses. This is important, as various subsets of features in $M$ may be chosen, or the features in $M$ can be used in different ways to provide an understanding of agent responses:
Scale and granularity directly affect how the agent's responses are perceived: For instance, considering scale, a trajectory may be considered in a space of $||M||$ or less dimensions, given different projections of the trajectory in subspaces. As a concrete example, consider  a trajectory in the 3D space, where $M=\{longitude, latitude, altitude\}$: One may view such a trajectory in any 2D projection, thus considering responses in any subset of features in $M$. Generally, we may consider the ``projection" of responses $p(\sigma^{}_E, S_M)$ in any  subset $S_M$ of $M$. Granularity may impose filtering and/or aggregation of features in the temporal dimension.
%, which should be can be also added to the denotation of such a response (E.g., $p(\tau_{E,H}^{\theta, \phi}, S_M)=\tau_{E,H,S_M}^{\theta, \phi}$). 
On the other hand, a critical aspect to be considered about features in $M$ is how the agent's responses are affected, if any state of $E$ is projected in a $S_M \subset M$?  This is in contrast to the simple response projection mentioned above, as it affects how the agent perceives the environment: Such ``masking" of features and ``perturbations" reveal causalities and sensitivities on environment features. 

The last question concerns whose needs does explainability addresses. As already pointed out, in this article, we focus on the needs of human operators, who require to solve  domain problems in real-life contexts, rather than on the needs of machine learning and RL experts. This is connected to several aspects considered above, such as 
(a) 
how humans' beliefs and understanding of agents' inner workings are modeled, 
(b) 
how and in which terms  humans are assumed to perceive and understand the DRL inner workings, 
(c) 
in which scale and granularity DRL responses are provided, 
(d)
how sensitivity on features and features' importance is communicated in specific circumstances and overall, and  
(e) 
how, specifically, the policy and/or the objective function are made understandable (also in connection to (b)).

In this article, we do not elaborate on  issues related to human factors and modelling of humans, nor on the effectiveness of different surface representations of explanations. There are many works exploring these issues and it is out of this article's scope. 

\subsection{Explainability desiderata}
\label{sec:Desiderata}
An agent, to be explainable, needs an interpretable model and an explanation logic. The interpretable model provides explanations' content, while the explanation logic selects the modes and the way explanations are surfaced, towards meeting transparency requirements. Interpretable models may be functional parts of the agent, i.e., the agent may have the ability to provide responses without them, as we will explain in subsequent sections.

At high levels of automation, the transparent agent must be   \textit{trustworthy}, in the sense that operators should be comfortable relinquishing control to it, as pointed out by Z. C. Lipton in \cite{Lipton}. A major factor to that is the \textit{accuracy of predictions and ``goodness" of prescribed actions}  in terms of solving the identified problems (predictive accuracy), w.r.t the interests of stakeholders (e.g., without increasing cost of operations or compromising safety). However this is not sufficient: Transparency is crucial, especially when system's responses do not comply with experts' usual practices or intuition, and in cases where operators need to understand system's capabilities and limitations.

There are different desiderata for transparency and explainability, regarding human interaction/communication issues, and aspects concerning properties of the interpretable models. This becomes apparent if we consider that transparency requires explainability that meets pragmatic constraints, which in its turn requires interpretable models that provide information  towards enhancing humans' understanding of systems' responses, objectives and capabilities. 

Thus, desiderata concern providing explanations that are (a) \textit{accurate on the information provided}, (b) \textit{ fitting the purpose and context of system deployment}, (c) with respect to  \textit{operational requirements} (e.g., on the type of information provided per type of problem instance), \textit{operational constraints} (e.g., time constraints and other constraints regarding privacy and security, etc.), and with respect to  \textit{ethical aspects}.

Towards satisfying these desiderata, the \textit{type and amount of information} being provided and the \textit{way explanations are being surfaced} are crucial.  Scale and granularity of responses are important aspects.
Specifically, concerning the type and amount of information provided, a system may provide a systematic trace of its inner-workings (low scale), with an exhaustive enumeration of causal paths and features being considered per problem instance (low granularity), increasing the complexity of explanations, which may also be affected by the size and complexity of the interpretable model. The ways in which this information is communicated affects transparency and trustworthiness: People need to accomplish tasks with awareness of the problem(s) faced, also with respect to pragmatic constraints and ethical issues (e.g., to sensitivity of information, to potential dual-use of information and system,  fairness - making no discrimination against groups of entities) and with respect to any contextual factor that affects acceptability of system's responses and explanations. For instance, a lengthy explanation regarding a system response to an emergent problem instance (with details known to the explainee), may lower explanations' acceptability and can determine system use in contexts where timely response is necessary, unacceptable. 
Selecting the most important causes/features that should be included in an explanation, focusing especially on those that are of interest to operators w.r.t the problem instance considered,  is very important, as evidenced by studies regarding human cognition and explanation effectiveness \cite{deGraafMalle}, \cite{Miller}. This supports effective update of  operators' models on agent's policy, objectives, preferences and capabilities, relating operators' existing knowledge and agents' interpretations.

\textit{Causality} and \textit{robustness} are additional features for the system to be trustworthy and these properties must be conveyed to humans via the explanations provided: While causality requires identifying the input features and causal chains that affect  system's responses, robustness requires that the system performance is not affected considerably with small variations of input features, especially of those that are not important to problem instances: Importance of features per problem instance, and sensitivity on features, are important elements for explanations. 

Regarding accuracy, we distinguish predictive accuracy, and the  accuracy of the information provided by the explanation logic (descriptive accuracy). While accuracy of predictions concerns the ML method used, accuracy of the information provided in explanations, concerns the capacity of the interpretable model and of the  explanation logic to locate and present the exact models' elements affecting agent's responses. 
In cases where the interpretable model is distinct from the models used by the decision-making agent, then the interpretable model must replicate agent's behaviour for explainability purposes with  \textit{fidelity}: The interpretable model must reproduce agent's responses accurately, identifying also the \textit{important features} that affect agent's responses.
While this may happen with \textit{local fidelity} (i.e., in the vicinity of the problem instance being solved), the interpretable model may provide \textit{global perspective} of the agent's behaviour (e.g., explaining objectives, intended outcomes, preferences, capabilities). These important distinctions are clarified by the problem definitions stated in the next section.

\subsection{Problem formulation}
\label{sec:Formulation}

Based on the goal of DRL agents, as specified in section \ref{sec:preliminaries}, and on the  explainability / transparency goals and desiderata, as specified in sections \ref{sec:interpretability} and \ref{sec:Desiderata}, respectively,  this section  provides definitions for the \emph{model explanation}, \emph{outcome explanation} and \emph{model inspection} problems for DRL methods, refining those provided in \cite{Survey}, so as  to connect explainability for DRL methods to explainability of ML prediction methods, allowing a more comprehensive view of explainability in ML: 
Overall, the model explanation problem aims at providing explanations regarding the overall, global  logic behind agents'  responses, revealing also global information on  agents' preferences and capabilities.  The outcome explanation problem is related to explaining responses provided by agents under certain circumstances, specifying the correlation between responses and values of features in $M$, at a low level of granularity (i.e., regarding certain actions) and scale (i.e., in restricted temporal and spatial extent). Finally, the model inspection  problem concerns the provision of information that reveals aspects and properties of the individual models that a DRL method comprises.

\emph{Definition 2.4.1 (Model Explanation Problem)}. Given a DRL method, the \emph{model explanation problem} consists in finding a global explanation $X \in \mathcal{X}$ of behaviour, belonging to a human interpretable domain $\mathcal{X}$ through an interpretable model $c_g(f(\theta), f(\phi); E)$, in a specific environment $E$, using $f(\cdot)$ that provides a representation of   policy and objectives models' elements. 
The global explanation $X$ is provided via an explanation logic $\epsilon_g(c_g)$ that reasons over $c_g$.

Providing a solution to the model explanation problem aims at updating $P(\sigma^{\theta, \phi}_E|f(\theta), f(\phi); E)$,  interpreting agent's responses at coarse levels of granularity and at large scale extents, exploiting agent preferences, abilities and objectives in $E$. We may say that solutions to the model explanation problem aim to provide global (i.e., independently from specific states)  explanations both for the policy and for the objectives of the agent. 

Refining Definition 2.4.1 to either the policy or to the objectives models of a DRL method, we have the \emph{policy model explanation problem} and the \emph{objectives model explanation problem}:

\emph{Definition 2.4.2 (Policy / Objectives Model Explanation Problem)}. Given a DRL model, the \emph{policy (objectives) model explanation problem} consists in finding an explanation $X \in \mathcal{X}$ of behaviour, belonging to a human interpretable domain $\mathcal{X}$ through an interpretable  model $c{_g}{_p}(f(\theta) [f(\phi)]; E)$ for policies ($c{_g}{_o}(f(\phi); E)$ for objectives), using a representation of agent's policy or objectives  models' elements, provided by $f(\cdot)$. 
The global policy/objectives explanation $X$ is provided via an explanation logic $\epsilon{_g}(c{_g}{_p})$ (respectively,  $\epsilon{_g}(c{_g}{_o})$) that reasons over $c{_g}{_p}$ (respectively, $c{_g}{_o}$).

It must be  noted that in the above definition, the policies' interpretable model $c{_g}{_p}(f(\theta) [f(\phi)]; E)$ includes optionally (this is what square brackets denote) a representation of agent's objectives: We leave this as an option, given that the objectives may be  assumed to be known, or interpreted separately. 

\emph{Definition 2.4.3 (Outcome Explanation Problem)}. Given a DRL method, the \emph{response explanation problem} consists in finding a local explanation $X \in \mathcal{X}$, belonging to a human interpretable domain $\mathcal{X}$ through an interpretable model $c_l(f(\theta), f(\phi), s_t ; E)$ at state $s_t$, using a representation of agent's policy and objectives  models' elements, provided by $f(\cdot)$. 
The local explanation $X$ is provided via an explanation logic $\epsilon_l(c_l, s_t)$ that reasons over $c_l$ and $s_t$.

It must be noted that the outcome  explanation problem concerns  skills and actions that the agent applies at states $s_t$, as well as (sub-)trajectories that at time $t$ are in state $s_t$.
In any case, the response explanation problem concerns agent's responses in fine granularity and small scale. Additionally, it may also concern providing local interpretation of the objective function, motivating some responses instead of others.

\emph{Definition 2.4.4 (Local Response/Objectives Explanation Problem)}. Given a DRL method, the \emph{local response/objectives explanation problem} consists in finding an explanation $X \in \mathcal{X}$, belonging to a human interpretable domain $\mathcal{X}$ through an interpretable  
model $c{_l}{_\sigma}(f(\theta) [f(\phi)], s_t; E)$ for responses (respectively, $c{_l}{_o}(f(\phi), s_t; E)$ for objectives) at state $s_t$, using representations of models' elements provided by $f(\cdot)$. 
The local explanation $X$ is provided via an explanation logic $\epsilon{_l}(c{_l}{_\sigma})$ (respectively,  $\epsilon{_l}(c{_l}{_o})$) that reasons over $c{_l}{_\sigma}$ (respectively, $c({_l}{_o})$).

%Consistently with the global response model, the local model includes optionally the process of representing agent objectives.

Providing a solution to the outcome explanation problem aims at updating a model $P(\sigma^{\theta, \phi}_E|f(\theta), f(\phi); E)$, towards  interpreting agent's  responses and objectives at fine levels of granularity and scale, at any state. We may say that any solution to the outcome explanation problem aims at providing local  joint explanations both for the fine grained responses and objectives of the agent, at particular states. 

The model inspection  problem for DRL methods concerns providing surface representations of DRL internal models' elements that would reveal agent's internal models' workings and properties, provided via the function $f(\cdot)$. 

\emph{Definition 2.4.4 (Model Inspection Problem)}. Given a DRL method, the \emph{model inspection problem} consists in finding a surface representation $r=R([f(\theta)]+[f(\phi)]; E)$  of DRL models' elements, or of models' properties, exploiting  representations of models' elements provided by $f(\cdot)$. 
The notation $[f(\theta)]+[f(\phi)]$ indicates that  $R$ exploits either $f(\theta), f(\phi)$, or both, in a joint manner.

In contrast to providing model explanations, the model inspection problem concentrates on the presentation of individual models' elements and properties at low levels of detail, towards inspecting the inner workings of models, without necessarily revealing an understanding of agent's policy, objectives, or responses. %Additionally, the problem here concerns the surface representations of models' parameters and/or input features, rather than affecting human understanding of these (e.g., via examples, or inspection of specific paradigmatic cases).

\section{Framework and paradigms for XDRL methods}
\label{sec:framework}

Given the above stated definitions of the DRL explainability problems, Figure 1 provides the overall architecture of a framework for XDRL methods:
In addition to the inner dark box, which provides the main components of any DRL method\footnote{This comprehensive DRL blueprint has been suggested by S.Levine, according to our knowledge.}, Figure 1 presents the basic pipeline for XDRL. DRL models provide interpretations, denoted by $c(\cdot)$, which, moving from dark to lighter areas of the blueprint, are exploited to provide content to an explanation logic. It must be noted that, as defined above,  any such model exploits either $f(\theta), f(\phi)$, or both, in a joint manner.

%Before delving into the details of the framework, it must be noted that this  blueprint can be instantiated with a number of components that provide interpretable models, realizations of $f(\cdot)$), and explanation logics: Figure 2 provides an instantiation with components for providing explanations on objectives, the policy and responses.

The policy $\pi_\theta$ of  a DRL agent  is represented by the  \emph{policy model} with parameters $\theta$ (e.g., a deep neural network (NN)). Given environment states' features in $M$, this model provides a probability distribution on all  actions that can be applied at a state. 
In conjunction to that model, there can be other models either fitting a value function (e.g., $V$ on states or $Q$ on state-action pairs), or modelling the way the environment evolves, or fitting the reward function. Figures 1 and 2  show only the reward model, placing special emphasis on the objectives of the agent. This model approximates the reward function $r_\phi$ at any time instance $t$, given a state-action pair $(s_t, a_t)$.   
Overall, the DRL method exploits the reward model (possibly, in conjunction to  other models) to fine-tune the policy model parameters $\theta$, while the agent may  generate trajectories' samples in $E$ to test the optimality of the policy w.r.t. the objectives.

\begin{table}
\small
    \centering
    \begin{tabular}{c c}
  \includegraphics[width=0.4\textwidth]{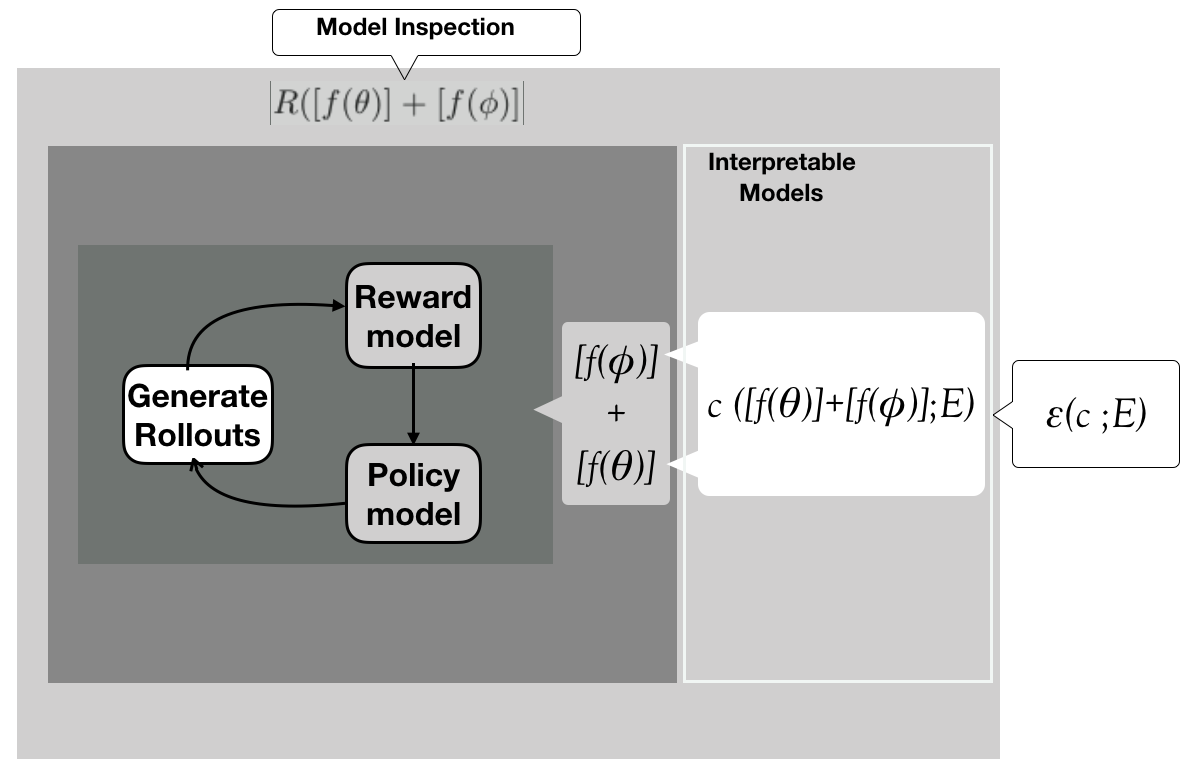}
     & 
  \includegraphics[width=0.4\textwidth]{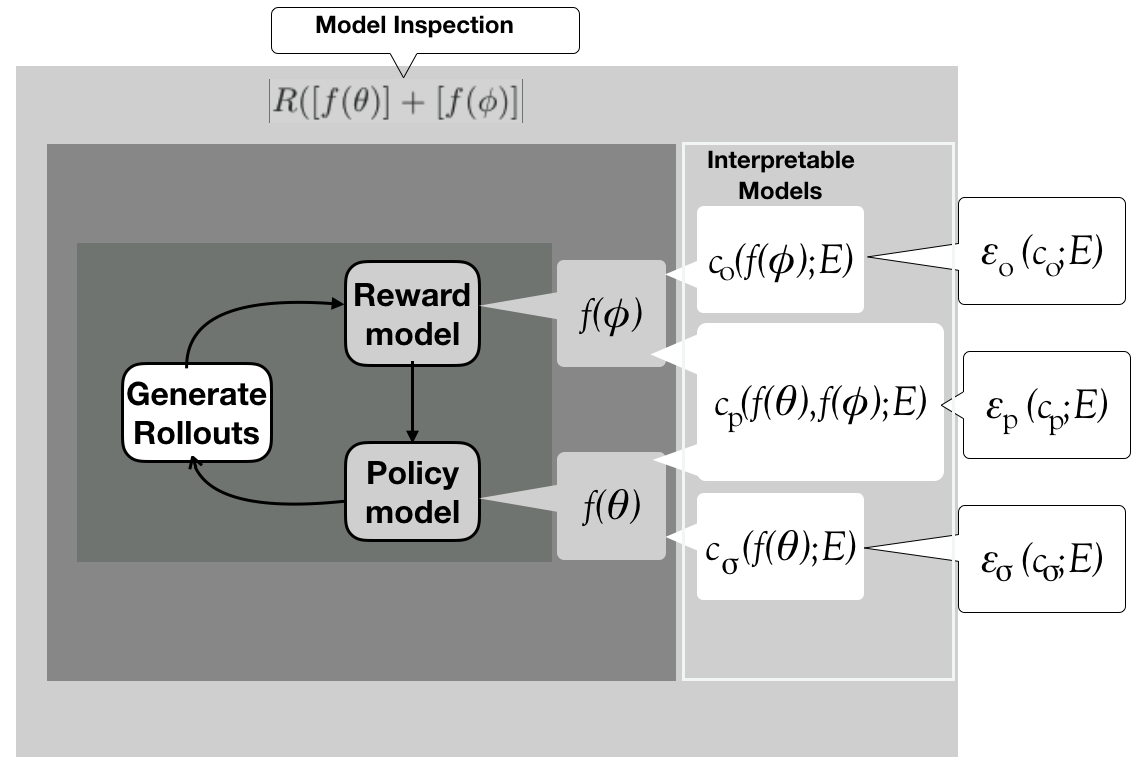}\\
  Figure 1. Overall framework for XDRL &
  Figure 2. Framework for XDRL \\
  &
  with  interepretable models

    \end{tabular}
%    \caption{Framework for XDRL: Overall architecture (left), With specific interpretable models (right)}
    \label{tab:figoverall}
\end{table}

To inspect individual models, function $f(\cdot)$ provides a representation of policy model and of objectives model elements. This may solve the model inspection problem, but it does not solve any of the (global policy or objectives) model  or  (local responses / objective) outcome explanation problems.

 Figure 2 shows a refined version of the XDRL framework with  interpretable models $c$ exploiting the representations provided by $f(\cdot)$. As it is shown, any such 
 interpretable  model may be an objectives model ($c_o(f(\phi);E)$), a local responses model ($c_\sigma(f(\theta), s_t ;E)$), or a  policy  model ($c_p(f(\theta), f(\phi);E)$). Any XDRL agent may contain all or any combination of these interpretable models, at the local and/or at the global level. %It must be noted that the global model $c_g$ may include the global objectives $c_g_o$ model in conjunction to the global policy model $c_g_p$, however, as Figure 2 shows, we distinguish clearly the objectives' models $c_o$. 

 Explanations are generated by functions $\epsilon$ that exploit the interpretable models, maybe in combination with the elements provided by the model inspection components. Figure 2  shows functions providing explanations for objectives ($\epsilon_o(c_o;E)$), responses ($\epsilon_\sigma(c_\sigma;E)$) or for the  policy model ($\epsilon_p(c_p;E)$).

\subsection{Interpretable methods}
\label{sec:transparent}

\begin{table}
%    \centering
\small
    \begin{tabular}{c c}
      \includegraphics[width=0.4\textwidth]{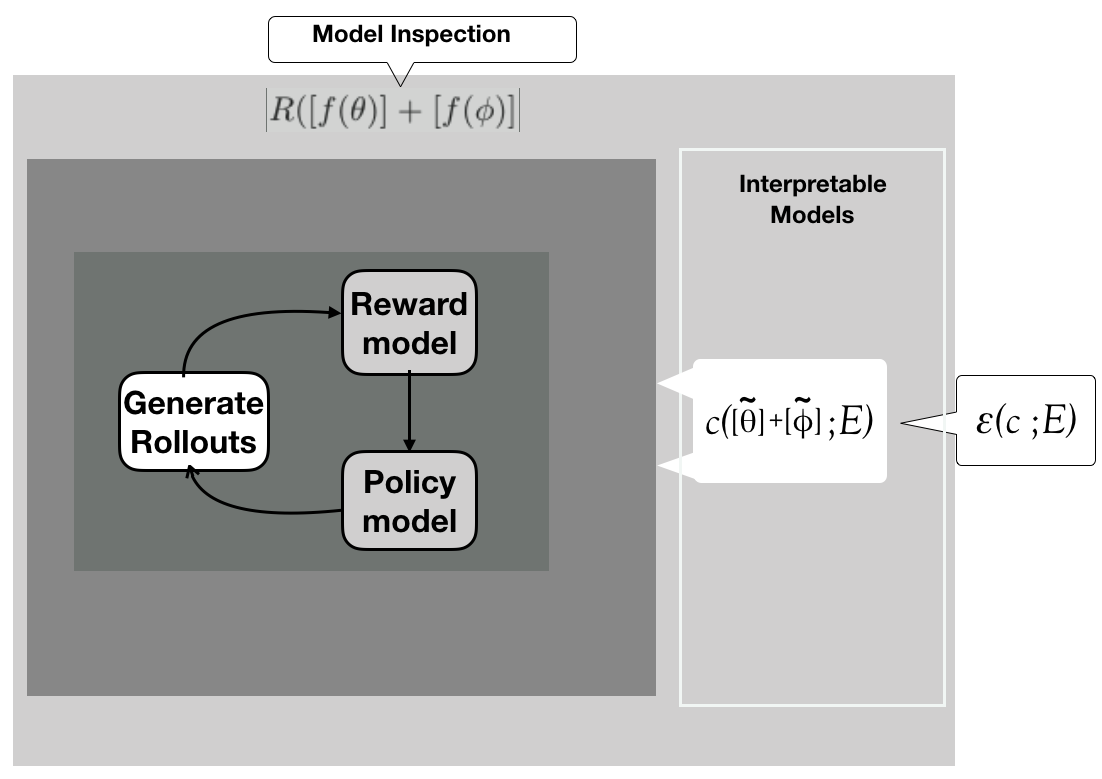}
         & 
        \includegraphics[width=0.4\textwidth]{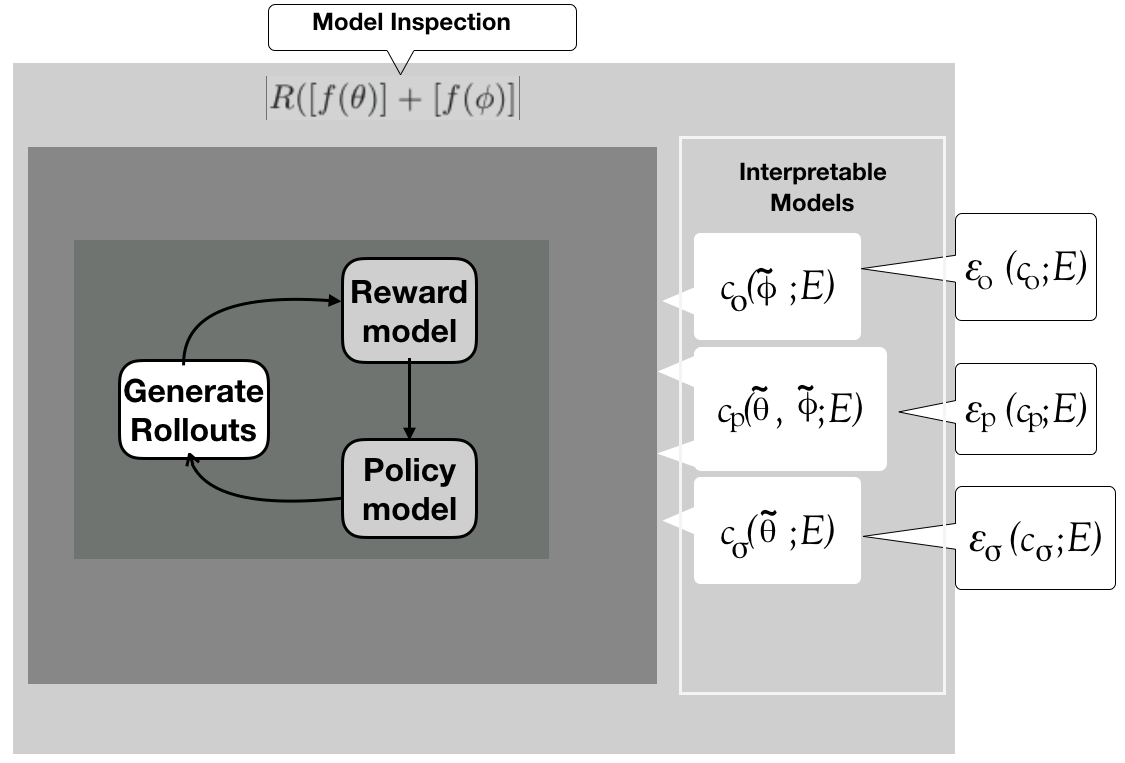}\\
        Figure 3. Overall framework for the interpretable box design paradigm &
        Figure 4. The interpretable box design paradigm\\
        & with specific interpretable models
    \end{tabular}
%    \caption{interpretable models for XDRL following the transparent box design paradigm: Overall architecture (left) with specific interpretable models (right)}
    \label{tab:transparent}
\end{table}

Some of the models that a DRL agent exploits may be directly interpretable: For instance, the policy model may be a decision tree, or the reward model may be a  linear function on features in $M$.  In this latter case, $\phi$ realizes the weights of the reward function 
$r_\phi(s_t,a_t; E)= \phi^\text{T} \mu(s_t, a_t; E)$,
where $\mu$ are functions on features in $M$ and action parameters. As   functions $\mu$ become arbitrarily complex, they hinder the interpretability of the reward model. %In some cases (e.g., in inverse learning methods) reward functions may also be fitted by (deep) non-linear models with parameters $\phi$.

Interesting variations of the XDRL framework provided in Figure 1, are frequently met in the literature, where interpretable models $c$ either  (a) substitute inner constituent models of the DRL  method (thus making the DRL method ``inherently" interpretable), or (b) are being trained using the DRL agent and its models as oracles in order to provide explanation content. These cases, in close relation to definitions provided in \cite{Survey}, aim to solve the \emph{transparent box design problem}. In this paper, these variations are considered to offer distinct paradigms for building interpretable models, 
according to what we call the \emph{interpretable box design paradigm}\footnote{We do not use the term ``transparent", as we associate transparency with pragmatic aspects regarding the provision of explanations.}. 

%A DRL agent being developed according to the interpretable box design paradigm uses interpretable models that either substitute or replicate inner agent's models. 
Interpretable models, as Figure 3 shows, are parameterized using variables' vectors $\tilde{\theta}$ for policies and $\tilde{\phi}$ for rewards (possibly in conjunction to other models), distinct from the parameters of the DRL models, if separate DRL models exist. It must be noted that interpretable models in these cases do not necessarily exploit DRL models' elements by means of $f(\cdot)$.
As a consequence, while model inspection facilities may be provided for the  constituent DRL models, these in some cases are totally disconnected from the interpretable models.

Interpretable models using the DRL method as an oracle (the second variation mentioned above), can be trained either during or after training the DRL models. %Being trained during the training of the RL models, the interpretable model evolves as the inner model evolves, and can be used to explain how the training process affects agent's responses. Finally, when a model is being trained  by a trained DRL method, it aims at constructing an interpretable model of any DRL constituent model.  
While training interpretable models  during training the DRL model may result to instability and inefficiency of the training process, aiming to reach a moving target, it may provide insights to the DRL learning process. The training process may use samples and results  (e.g., rewards, state/action values) provided by the deep method, while it may also exploit the models of the DRL method in a direct manner, delving into their inner workings. 
This results into two distinct paradigms that refine the interpretable box paradigm: The \emph{mimicking} and the \emph{distillation} paradigms, respectively.

It must be noted that although the distillation process has been introduced as a NN model compression process through training a student NN, and has been introduced as a term in \cite{HintonDistillation}, the first work that introduces it for sequential decision making (as a model compression technique, not as a method for building interpretable models) is described in reference \cite{PolicyDistillation}, where the teacher  provides samples, each comprising an observation sequence and a vector of state-action $Q$-values. 

For DRL methods, the distinguishing line between the distillation and the mimicking processes is not that clear in the literature: Here we consider these as two distinct paradigms for providing interpretability to DRL methods.

\subsection{Distillation and mimicking: Two distinct paradigms for interpretability}
\label{sec:paradigms}

\begin{table}
    \centering
    \small
    \begin{tabular}{c c}
      \includegraphics[width=0.4\textwidth]{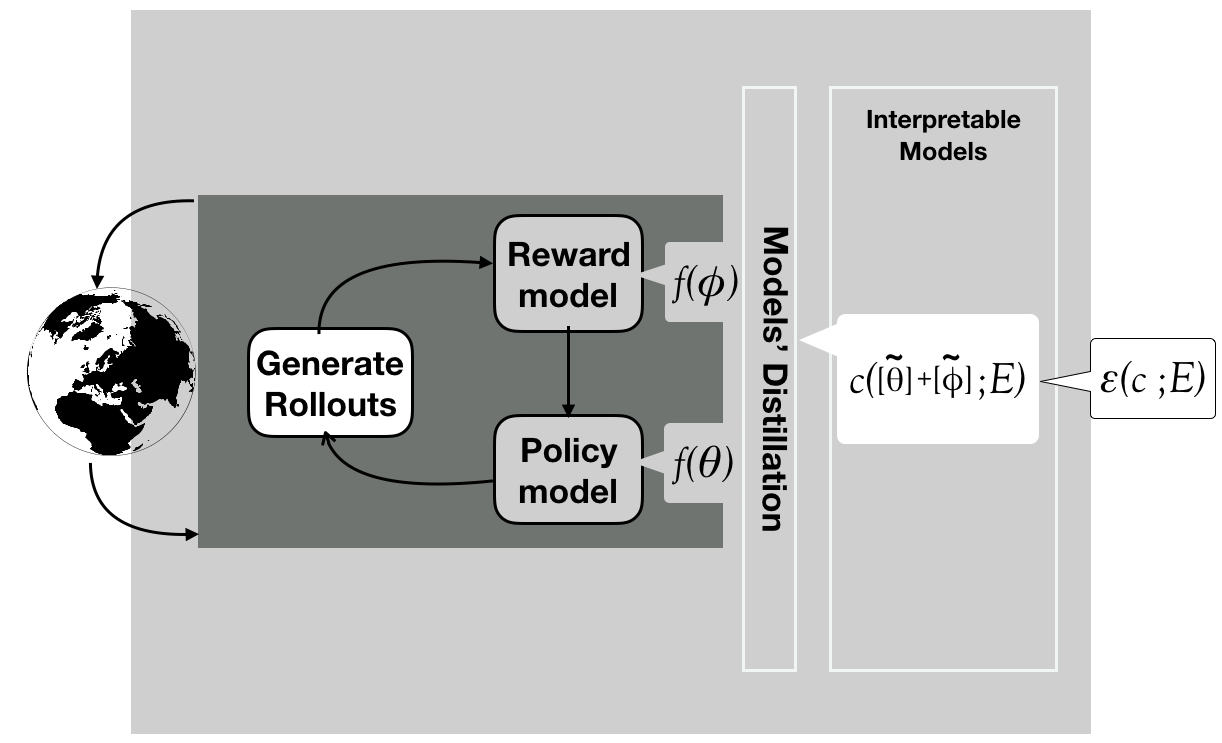}
         & 
        \includegraphics[width=0.4\textwidth]{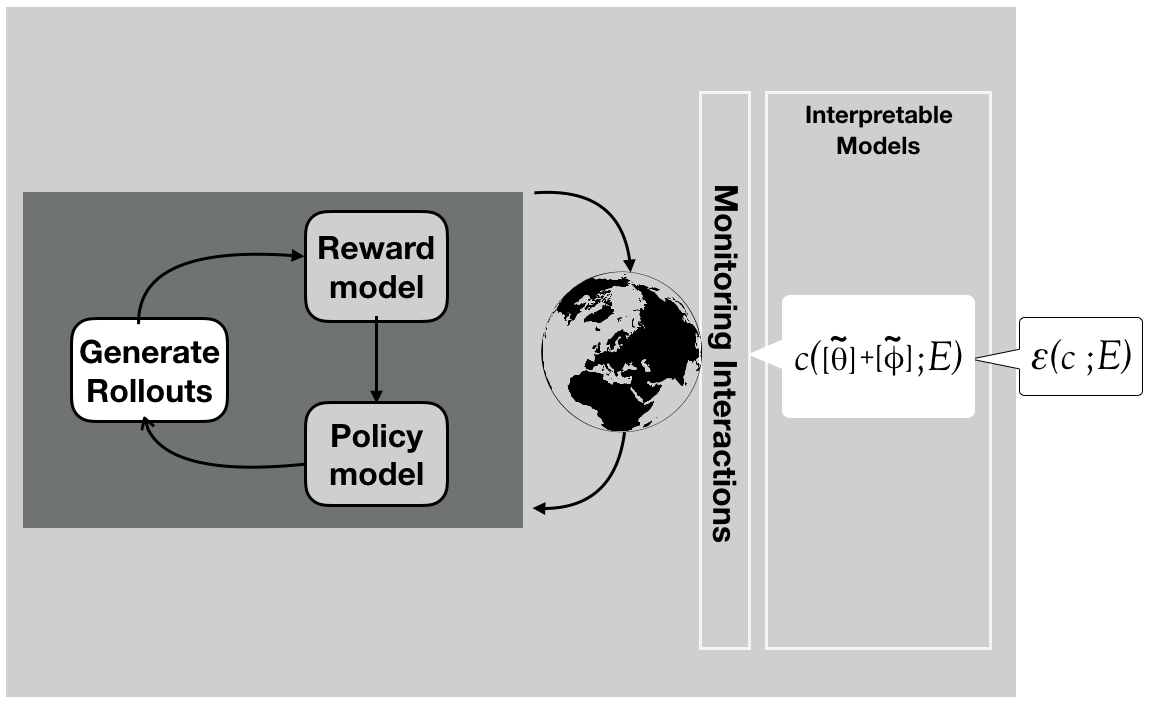}\\
        Figure 5. The overall framework for distillation  &  Figure 6. The overall framework for mimicking
    \end{tabular}
 %   \caption{Producing interpretable models for XDRL via distillation  (left) and mimicking (right)}
    \label{tab:distill-mimick}
\end{table}

The distinct paradigms are shown in Figures 5 and 6. The main difference between the two paradigms concerns the ``receptive field" of the interpretation process:
As Figure 5 shows, the distillation process gathers knowledge from the trained DRL agent, exploiting  any of the constituent DRL models {\em in a direct way} through a process of transforming DRL models' elements into interpretable models.
In contrast, the mimicking process, as Figure 6 shows, monitors the interaction of the  RL agent with the environment, and gathers interaction samples, recording agent decisions, state transitions, rewards and consulting DRL assessed values.

As an important note to the above, one may argue that the architecture shown in Figure 5 is more general than the one shown in Figure 1. Actually,  the architecture shown in Figure 1 abstracts all XDRL paradigms while providing \emph{exact} interpretations of DRL models. 
The two paradigms offer interpretable models that are approximations of the DRL constituent models: The distillation paradigm shown in Figure 5 aims at exploiting the knowledge acquired by the DRL models via model inspection facilities, while the mimicking method in Figure 6 produces interpretable models  without  exploiting the inner working components of DRL agents.  

\section{XDRL methods: State of the Art}
\label{sec:stoA}

This section aims at a detailed presentation of  XDRL approaches. For each of the reviewed approaches, we describe the overall approach and motivating points, assumptions and important technical details, and we conclude with important aspects w.r.t. the specific problem it addresses, paradigm it follows, interpretable models it constructs, required access to constituent closed-box models, and finally, information on evaluation objectives, methods and results.
In addition to these, the Appendix  summarizes the XDRL approaches described, focusing on important aspects of their functionality and applicability, using a comprehensive list of characteristics that includes those described above, and additionally, the XDRL requirements for  the DRL methods used, as well as ``XDRL - DRL interplay aspects": Access to DRL models' information and/or samples required, in an online mode or in an offline mode.

As a first note before delving into XDRL methods, there are many  proposals for interpreting deep NNs models, mainly through  distillation and mimicking approaches. The main emphasis of these approaches is building high-fidelity interpretable models, in comparison to the original models. These approaches differ in several dimensions: (a) the targeted representation (e.g., decision trees in DecText \cite{ExtractingDTfromTrainedNNs}, logistic model trees (LMTs) in reference \cite{LMTExtractionfromANNs}, or Gradient Boosting Trees in reference \cite{interpretableDeepModelsforICU}), (b) to the different splitting rules used towards learning a comprehensive representation, (c) to the actual method used for building the interpretable model (e.g., \cite{LMTExtractionfromANNs} uses the LogiBoost method, reference \cite{ExtractingDTfromTrainedNNs} proposes the DecText method, while the approach proposed in reference \cite{interpretableDeepModelsforICU} proposes a pipeline with an external classifier, (d) on the way of generating samples to expand the training dataset. 
These methods  can be used towards interpreting constituent individual DRL models employing (deep) NNs. We do not aim to  review these in this article, given the thorough review provided elsewhere  (e.g. in \cite{Survey}, \cite{Murdoch22071}, \cite{PrinciplesPracticeofXML}, \cite{InterpretableML}). However we delve into  XDRL approaches that exploit any of these methods.

\subsection{Solving  the model inspection problem}

Here we  provide a succinct review of prominent general methods addressing the DRL model inspection problem.

\subsubsection{Why should I trust you?}
\label{sec:4.1.1-LIME}

M. T. Ribeiro, S. Singh and C. Guestrin in \cite{LIME} propose LIME for explaining the predictions of any classifier in an interpretable and faithful manner. LIME learns an interpretable
model locally around the prediction, searching for instances in the vicinity of the original instance through perturbations  of uniformly sampled features in subsets of $M$ (i.e., non-zero elements). 
Emphasis is given to the qualitative understanding of correlations between input features and model responses, aiming to guarantee local fidelity, explaining how the model behaves in the vicinity of the instance being predicted, using a proximity measure.
Thus, while LIME addresses the model inspection problem assuming binary vectors for instances' interpretable representations, it  contributes to the outcome explanation problem solution, as it provides the importance of features in a local only context. 

Building on explanations of individual predictions, authors in \cite{LIME} propose the SP-LIME
method by presenting representative individual predictions and their explanations in a non-redundant
way, framing the task as a sub-modular optimization problem. SP-LIME provides a  global perspective and contributes to solving the model explanation problem. 

Experiments performed with simulated users regard (a) the faithfulness of explanations to the model, given a set of gold features that explanations must recover; (b) the trust on predictions, given untrustworthy features in predictions; (c) the trust on the model, given a set of trustworthy examples covering cases with varying subsets of important features.  Evaluations with human subjects regard the accuracy of choosing the best classifier, and improving the classifier by eliminating features which are untrustworthy (measuring classification accuracy after rounds of interaction).

LIME provides a generic (model-agnostic) method that contributes to local and global (model) interpretability. In the context of DRL, it can be used for explaining any of the models (e.g., to explain local decisions for selecting an action given an interpretable representation of states' features), although not being designed for that particular task. Functions $f(\cdot)$ provide binary vectors for instances' representations, while $c_l$ (for outcome explanation ) can theoretically be any interpretable model, e.g., a linear model, a decision tree, etc. %LIME provides an explanation on the classification of any specific instance via features' perturbations. 
Surface representations of instances' interpretable representations (for model inspection, via $R$), as well as explanation logic $\epsilon_l$ for local explainability  and $\epsilon_g$  for model explainability, are provided using visual means. 

Extensions of LIME include Anchor \cite{ANCHOR} that uses decision rules as local interpretable classifier $c_l$, and LORE \cite{LORE} that learns a local interpretable predictor $c_l$  on a synthetic neighborhood using a genetic algorithm.

\subsubsection{Shapley Additive Explanations}
\label{sec:4.1.2-SHAP}

The Shapley Additive Explanations (SHAP) approach aims to provide a unified approach to interpreting model predictions, showing how interpretation methods are related, and when one method is preferable over
another. S. M. Lundberg and L. Su-In in \cite{SHAP} identify the class of additive feature attribution methods with a linear function of binary variables as an explanation model:
$g(z)=\phi_0+\sum_{i=1}^{SIF}\phi_iz_i$,
\noindent where $z \in \{0,1\}^{SIF}$, $SIF$ the number of simplified input features and $\phi_i \in R$.

There is a family of methods - including LIME, DeepLIFT \cite{DEEPLIFT}, Layer-Wise Relevance Propagation \cite{LayerWiseRelPropagation} and classic Shapley value estimation methods- that are additive feature attribution, and authors in \cite{SHAP} show to unify six of the existing methods. They provide
theoretical results on the existence of a unique solution in this class of methods with a set of
desirable properties, that
several recent methods in the class lack:  \emph{Local accuracy}, requiring the explanation model to at least match  the output of the original model; \emph{missingness}, representing features presence so as to locate the features with no prediction impact on a local decision; and \emph{consistency}, regarding the  relation of input's contribution and input's attribution.  

Specifically, considering a simplified input $x'$  mapping to the original input $x$ through the
mapping function $h_x$: $x = h_x(x')$, and input $z' \approx x'$, with $S$ to be the set of non-zero indexes in $z'$ ($z_S$ has missing values for features not in $S$) 
s.t. $h_x(z')=z_S$, SHAP values are the Shapley values of a conditional expectation function of the original model $E[pred(z)|z_S]$, given the model prediction $pred(z)$.
SHAP focuses on explaining a prediction $pred(x)$, given a model $pred$ with input $x$ and a simplified input $x'$. Such local methods ensure that $g(z') \approx pred(h_x(z'))$, whenever $z' \approx x'$, where $g$ is  the local interpretable model $c_l$ (for outcome explanation).
%As such, SHAP can be used for interpreting any of the constituent DRL models realized as predictors, but it has not being designed for solving the DRL model nor the outcome explanation problem for sequences of decisions, at any granularity level and scale.
SHAP, similarly to LIME, provides an interpretable model as a linear function of binary vectors for instances' representations provided by $f(\cdot)$. Surface representations $R$ of instances' interpretable representations for model inspection, as well as explanation logic $\epsilon_l$ for local explainability, are provided using visual means. 

%- Again it is used mostly for classification, but evaluated using dense and sparse decision trees,

Based on insights from unifying methods, authors in  \cite{SHAP} present methods that show improved computational performance and/or better consistency with human intuition than previous approaches.
Evaluation  of Kernel SHAP vs. LIME
and Shapley sampling values show improvement in terms of learning accurate estimates with fewer evaluations of the original model (sample efficiency). User studies in a  limited setting show that SHAP values prove more consistent with human intuition than alternative feature importance allocations represented by DeepLIFT and LIME, which fail to meet the desirable
properties identified.
 Finally, authors use MNIST digit image classification using a convolutional network to compare SHAP with
DeepLIFT and LIME, showing that SHAP is able to explain differences in classifying a digit to a class rather than to a near-class, measuring the probability of classifying instances to classes and changes in log-odds when masking over images.

A concrete example of employing SHAP (actually, a specific instance of LIME, named Kernel SHAP) in the context of DRL, is provided in reference \cite{XRL4TrafficSignals}.
This article uses Policy Gradient to implement a traffic signal control agent on a signalized
roundabout. %It analyzes the relation between the agent phase preferences and the actual traffic, assessing the agent capability of reacting to the current detectors state. 
Kernel SHAP estimates the effect of the road
detectors state on the agent selected phases, interpreting the decision taken, in relation to the traffic volumes and the
lanes occupancy.
\\ 

Other methods, such as for instance Layer-wise Relevance Propagation (LRP) \cite{LayerWiseRelPropagation} and Neighborhood Component Analysis (NCA) \cite{NCA} can be used for addressing the model inspection problem for any of the constituent DRL models, also providing (a) importance of features in $M$, and (b) reduction of predictors' dimensionality, which can facilitate data visualization and fast classification. Such methods are thoroughly reviewed in \cite{Survey}.

Following this short presentation of model inspection general methods, we proceed to review  XDRL methods addressing any of the model, outcome explanation, or the interpretable box design problems.

\subsection{Solving the policy explanation problem}

\subsubsection{ Summarizing agent behaviour to people with HIGHLIGHTS}
\label{sec:4.2.1-Highlights}

With the goal of increasing people's familiarity with agent's capabilities and limitations, D. Amir and O.  Amir in \cite{Highlights} propose the HIGHLIGHTS and the HIGHLIGHTS-DIV algorithms that choose trajectories  highlighting important and distinct aspects of agent behaviour. In doing so, this approach provides an overview of an agent's policy, rather than explaining specific responses.

Specifically, the HIGHLIGHTS online algorithm constructs summaries of agent's behaviour by selecting (sub-) trajectories with important states. A state is considered important if taking a ``wrong" action in that state (i.e., an action that is not prescribed by the policy) can lead to significant decrease in future rewards. To provide context to the user, for each important state the algorithms extracts a sub-trajectory including neighboring states (before and after the important states) and actions. In order to highlight behaviour in various sub-spaces in the state space, HIGHLIGHTS-DIV avoids including in the summary (sub-) trajectories with very similar  states, if these are equally important.

HIGHLIGHTS has been evaluated by playing the Pacman game, based on the observed behaviour rather than scores. Q-values in this setting are defined as a weighted function of state feature values. Given different agents trained for different number of episodes, human participants were asked to select the best agent based on the summaries provided, while subjective opinions about the helpfulness of summaries were elicited, given summaries of the same agent. The surface representation of the explanation in \cite{Highlights} is provided by means of video clips, but in general, other means and information modalities may be used (e.g., as in \cite{ContrastiveExplanations} that is presented in a subsequent section). This evaluation resulted in observations regarding participants confidence in choosing agents  based on HIGHLIGHTS summaries. Experiments show that short summaries may not enable participants to correlate their confidence with their ability to assess agent's abilities. However, with statistical significance and for very well trained agents, participants preferred summaries generated by HIGHLIGHTS over baselines, and summaries generated by HIGHLIGHTS-DIV over HIGHLIGHTS. 

Overall, this approach addresses5 the policy model explanation problem, demonstrating agent behaviour. It is important to point out that this approach by-passes the construction of an interpretable model $c$ for policies, but it  gathers explanations' content by selecting examples of agent's behaviour in important states. The representation of importance is realized via the function $f(\cdot)$ which has access to agent's policy and Q-values. 

T. Huber et al. in \cite{LocalAndGlobal} use a version of HIGHLIGHT-DIV in a combination of global and local explanations for DRL agents. Specifically, they use a version of the Layer-wise Relevance Propagation (LRP) \cite{LayerWiseRelPropagation} method, i.e. LRP with the {\em argmax} rule \cite{LRP-argmax}, to augment trajectory summaries with saliency maps.  This local approach is  presented in detail in section \ref{sec:4.4.11-LRP}. LRP has been  used to show what information is important to the agent towards a specific decision:  The approach presented in \cite{LocalAndGlobal} addresses the policy and the responses explanation problems by combining methods, contributing to our understanding of the role of saliency maps in the context of explaining agents' behaviour: While saliency maps have been shown to improve classification decisions in images (albeit with 60\% of correctness in \cite{Alqaraawi4}), their role in RL is not that clear (e.g. as shown in \cite{TransparencyandExplanationInDRL}). However, they may provide significant positive effects when combined with other methods (as in \cite{Erwig23}). Empirical evaluation of the proposed combined method aims to investigate (a) the mental model of humans about the DRL agent (via a retrospection task), (b) humans' ability to assess agents performance (via an agent comparison task) and (c) participants satisfaction with respect to the  explanations presented (via answering explanation satisfaction questions). The results of this study reinforce prior findings for HIGHLIGHTS-DIV. 
Overall, in this study, the choice of states shown to humans was more important than the inclusion of local explanations in the form of saliency maps. For saliency maps, the results reported are mixed: 
There were no significant differences between the saliency and nonsaliency conditions in the study. Showing saliency maps as part of a video is more challenging for humans, in comparison to showing them in a still image. As authors point out ``[saliency maps] in RL [add a] layer of complexity as interpretation also requires making inferences regarding how the highlighted regions affect the agent's long-term, sequential decision-making policy". However, in combination with behaviour explanations, saliency maps put the  important features in context, while they are not preferable when humans get highlights of agents' behaviour.

Extending the HIGHLIGHTS algorithms, P. Sequeira and M. Gervasio in \cite{InterestingnessElements} propose using not only state importance, but various interestingness elements so as to highlight important agent's behavioural aspects, increasing humans' understanding of agent's capabilities and limitations. Such elements are (in)frequent situations, (un)certain executions, observation minima and maxima, and most likely sequences to maxima, each serving a specific explanation provision purpose.  To compute these elements, the proposed framework collects data from agent's interaction with the environment, such as the number of times the agent has specific observations, estimated transition probabilities, as well as estimated Q-values and V-values.
Although there are elements (e.g., frequency) that may provide a good understanding of agent's behaviour characteristics, overall, results show that no single summarization technique — and hence
no single interestingness element — provides a complete understanding of every
agent in all possible situations of a task. Ultimately, different
combinations may be required for agents with distinct capabilities and
levels of performance. 

\subsubsection{Contrastive explanations for reinforcement learning in terms of expected consequences.}
\label{sec:4.2.2-ContrastiveExplanations}

Motivated by findings indicating that contrastive explanations offer intuitive motivations for humans to understand why one performs an action instead of another \cite{Miller}, J. van der Waa et al. in  \cite{ContrastiveExplanations} propose a method for providing contrasting descriptions of rollouts generated by an  RL agent's policy and a policy that is driven by humans' (non-experts in RL) queries. 

This proposal (a) defines a  translation of states and actions in a set of descriptive states' classes and action outcomes, by training binary classifiers, similarly to \cite{AutonomousPolicyExplanation}, and (b) proposes a method for obtaining a foil policy $\pi_f$ based on the foil in the user's query, suggesting potential alternative actions in a particular state $s$ and consecutive state-action pairs. This foil policy, based on an update of the $Q$ function to reflect user's preferences (and an informed update of the reward function) is contrasted to the fact policy $\pi_t$ learned by the agent. 
The explanations are based on generating rollouts of specific length simulating 
the effects of  $\pi_t$ and of $\pi_f$, based on the environment's transition function and starting on a state before the state specified in the user's query. State-action pairs in rollouts are transformed in descriptions which are then used for generating the trajectory contrastive descriptions by instantiating natural language templates.

Authors in \cite{AutonomousPolicyExplanation} report on a  case study using a simple RL setting with a very restricted objective, addressing either the next action of the agent policy or the entire policy: They show that human participants prefer explanations using the policy,  as in most cases the next action is evident to the user.

This approach addresses the policy model explanation problem, by contrasting  rollouts of the agent's policy to foil policy rollouts. The interpretable model $c$ provides  qualitative descriptions of rollouts by means of distinct state and action classes. The interpretation process accesses the MDP, which may be fitted by the learning process,  the state-action Q-values learned, and the rewards received, via the function $f(\cdot)$. Finally, the explanation logic $\epsilon$ uses natural language templates that are instantiated by qualitative  (potentially, contrasting) descriptions of rollouts.

\subsubsection{ Establishing appropriate trust via critical states.}
\label{sec:4.2.3-CriticalStates}

Motivated towards helping end-users to build a mental model of agent's policies and capabilities, S. H. Huang et al. in \cite{CriticalStates} propose approaches for computing critical states in a policy-agnostic way, requiring only access to action-value function $Q$. This proposal is driven by the insight that for many tasks the essence of the
policy is captured by agent's reactions in a few critical states.

A critical state $s$ is one in which it is
very important to take a certain action: As defined in \cite{CriticalStates}, $Q^\pi(s,a)$ varies greatly across different actions $a$, i.e., $Q^\pi(s,a)$  is very high for some actions, but for most of them, this value is mediocre or low. 
Given agent's policy, critical states are exposed to  end-users who can spot false-positive or false-negative critical states (i.e., states that are considered as critical by the agent, but not by the humans and vice versa). Inspecting the critical states, the end-users may refine their trust to the agent, or gain proper knowledge on the situations where they have to take control. For instance, both false-negative
and false-positive critical states indicate that the robot has
failed to learn something fundamental about the task. Similarly, if the policy
identifies a true-positive critical state but is mistaken about
which action is correct in that state, then the end-user may not trust the policy. 

This approach does not provide an explicit interpretable model but aims to provide a coherent view of agent's capabilities and limitations so as to establish trust to the agent. Critical states are determined by accessing $Q$-values via the function $f(\cdot)$. The  interaction with human experts concerns (a) spotting false-positive, false-negative states, or incorrect-actions in critical states and assessing whether to deploy the agent, (b) deploying the agent which operates with the user in the loop, who she is able to take control whenever needed. A critical issue is whether observing
a set of critical states leads participants to develop
appropriate trust in high-entropy policies that generated those critical states,  using the Soft Actor Critic algorithm (SAC). Also, while we may consider that this approach addresses the policy model explanation problem, it may be hard for the human to generalize and consider how the agent reacts in states  that have not been presented as critical. Also, while the surface representation of states-action pairs is shown to be depicted using visual means, it is harder to do so in settings where the dimensionality of the state/action space is large.

%%%%%%%%%%%%%%%%%%%%%%%%%%%%
%\subsubsection{Designing transparent boxes}
%%%%%%%%%%%%%%%%%%%%%%%%%%%%

\subsubsection{Graying the Black Box: Understanding DQNs}
\label{sec:4.2.4-GrayingBlackBox}

The main observation driving the work of T. Zahavy, N.B. Zrihem and S. Mannor in \cite{GrayingBlackBox}  is that Deep Q-networks (DQN) is learning an internal hierarchical model of the domain without explicitly being trained to. Therefore, authors propose a methodology and tools to analyze DQN, aiming to  identify spatio-temporal abstractions directly from the learned representation in two
different ways: First, manually clustering the state space using hand crafted features, and second, computing the Semi-Aggregated Markov Decision Process (SAMDP), an approximation of the true MDP learned from data, by means of spatio-temporal abstractions. Visualization tools, including t-SNE that exploits directly the collected neural activations of states (a  model inspection task), saliency maps, as well as  analysis tools for understanding the common attributes of states' clusters, and for analyzing dynamics between clusters, aim to provide facilities towards understanding, debugging and interpreting the policy model.

Specifically, authors in \cite{GrayingBlackBox} explain the five stages towards building an SAMDP model aggregating states and skills, thus allowing analysis with spatio-temporal abstractions: (0) Feature selection; (1) State aggregation, mapping the MDP feature space to the abstracted MDP state space enforcing temporal coherency among trajectories; %by considering all states relative to a state, in a time window $w$,
(2) Identification of skills (a.k.a sub-trajectories) from the data; (3) Inference of skill length, the SAMDP reward and the SAMDP transition probabilities; and finally, (4) selecting the best among the SAMDP candidates, according to evaluation criteria proposed.

Overall, this approach proposes a methodology and abstraction methods addressing  the policy model  explanation problem for DRL methods, where the policy is learned by the DQN method. The interpretable model $c_g$ is an SAMDP that is devised by consulting  manually crafted state features, as well as rewards and neural activations of states provided by $f(\cdot)$. The explanation logic $\epsilon_g$ is implemented by t-SNE maps with representative states per cluster, and appropriate t-SNE enabled SAMDP visualizations. In conjunction to that, authors demonstrate how to do (policy) model inspection by processing the network's neural activity using visualizations enabled by t-SNE.

%Up to now: Critical states and highlights, as well as state/actions abstractions

\subsubsection{Towards Better Interpretability in Deep Q-Networks}
\label{sec:4.2.5-betterInterpretability}

R. M. Annasamy and K. Sycara in  \cite{betterInterpretability} propose
an interpretable NN architecture for $Q$-learning,
which provides explanations for visual agents. The aim is to provide an understanding of the policy model, visualizing clusters of state representations in attention maps, and using saliency maps.

The proposed model learns a latent representation of states that captures important visual aspects of input images, and an association between representations of states and representations of keys in a key-value store using an attention mechanism.  Each key represents an action and an associated $Q$-value, among $N$ such values: Thus, the network learns to associate states to ``representatives" of (action,$Q$-value) pairs, which serve as cluster "exemplars". 
The attention weights over keys and their corresponding value pairs are used to calculate $Q$-values. Different types of losses are being linearly combined to train the network: Bellman error, distributive Bellman error, reconstruction error and diversity (over keys) error.  Uncertainty of the attention weights is being used to drive exploration during training, approximating an upper confidence on the $Q$-values.
Through key inversion (which not a process driven by a certain input) authors attempt to find important aspects of the input space
that influence the choice of a particular action-return
pair.

As concluded in \cite{betterInterpretability}, with a directed exploration strategy, the proposed model can
reach training rewards comparable to the state-of-the-art deep
$Q$-learning models. Specifically, the  best agreement between
the actions selected using the distributions in the image space
and latent space is rather low, but the explanations provide useful insight into the kinds of features extracted by convolutional layers: This possibly suggests, according to the authors, that reconstructions rely heavily on generalizability to unseen keys. However, results suggest that the features
extracted by the NN are shallow, and the agent does not model interactions between objects. 

Concluding, this approach addresses  the interpretable box design problem for deep $Q$-learning methods, focusing on providing global policy explanations for visual agents. The interpretable model $c_g$ is an attention model associating state embeddings to key-value embeddings. The explanation logic $\epsilon_g$ provides  keys - state embeddings clusters' visualizations using t-SNE for any specific $Q$-value, as well as  images reconstructed via key-value (i.e., action, $Q$-value) pair inversion.

\subsubsection{Interpretable Off-Policy Evaluation by Highlighting Influential Transitions}
\label{sec:4.2.6-InterpretableOPE} 
Off policy evaluation (OPE) estimates the value of a policy using data collected under another policy. Aiming to enable human experts to assess and analyze the validity of off-policy evaluation estimates, O. Gottesman et al.,  in \cite{InterpretableOPE} propose a hybrid human-AI system that highlights observations in the data whose removal will have a large effect on the OPE estimate. Their proposal aims (a) to address the needs of real-world applications where the data itself might never be enough; and (b) to involve domain experts in the integration of decision support tools, incorporating their expertise into the evaluation process. 
In so doing, they formulate a set of rules for choosing which observations to present to domain experts for validation.

Specifically, in this article authors propose a framework for using influence functions to interpret OPE,  and discuss the types of questions which can be shared with domain experts to use their expertise in debugging OPE: Influence functions show the impact that the removal of a transition has on state-action values or on the value of the evaluation policy. Based on this framework, they develop computationally efficient algorithms to compute the exact influence functions for several importance sampling estimators, as well as two broad function classes for Fitted Q-Evaluation (FQE): kernel-based functions and linear functions.

The  benefits of influence analysis  are demonstrated on a cancer simulator, with the objective to identify limitations in the evaluation process and make evaluation more robust. Furthermore, authors present results of analysis together with practicing clinicians of OPE for management of acute hypotension from a real intensive care unit (ICU) dataset.

Overall, this approach proposes a framework that  addresses off-policy evaluation independently from any reinforcement learning method. It is closer to  the policy explanation problem, since the influence analysis proposed aims to interpret a policy evaluation method. The method can be applied to parametric and non-parametric fitted Q-evaluation models, regarding continue or discrete states and actions, although the presentation and evaluation considers discrete cases. The method focuses on the influence of specific state transitions to the value of the policy, and thus, it can be considered to be a method that reveals the contributions of local decisions to the trajectories generated by a policy model. The model exploits trajectories generated by the evaluation policy, while there is not any specific proposal for the explanation logic: Different examples, use different visualizations of transitions’ influences.

\subsubsection{Interpretable Reinforcement Learning Using Attention Augmented Agents}
\label{sec:4.2.7-AttAugmentedAgents}

A.Mott et  al. in \cite{AttAugmentedAgents} propose a spatial attention mechanism to visual information in an
RL setting. The model enables  agents to actively select important,
task-relevant information from visual inputs, by sequentially querying and receiving compressed,
query-dependent summaries. Actually, the model generates
attention maps which can uncover the underlying decision process. Exploiting the attention maps one can understand how the system solves a task, identifying the type of entities that the agent attends (``what") and  the spatial locations where the agent focuses attention (``where").

Succinctly, the proposed model works as follows: Observations pass through a recurrent vision core
network, producing a ``keys” and a ``values” tensor, to both of which a spatial basis
tensor (encoding information about spatial positions) is concatenated. A query network  produces a set of query vectors taking as input the state of an LSTM recurrent network (policy network) from the previous time-step. The softmax of the  pixel-wise inner product between each query vector and each location in the keys tensor produces an attention map for the query. The attention map is broadcast along
the channel dimension, and it is point-wise multiplied with the values tensor. The result is then summed
across space to produce an answer vector. This answer is sent to the LSTM to produce
the next LSTM state.

The performance of the proposed method is evaluated  across
a broad range of ATARI levels. Attention maps are
used to visualize which parts of the input are attended. Authors conjecture that the top-down nature of
the attention (i.e., the fact that the attention map is produced by taking as input the LSTM states) provides a large performance gain compared to equivalent, bottom-up attention-based
mechanisms. Comparison of the attention maps
to alternate methods for visualizing saliency shows that they allow more comprehensive
analysis of the information the agent is using to inform its policy. The agent is able to make use of a combination of “what" and “where"
queries to select both regions and entities within the input, depending on the task. Experiments show that the agents are able to learn to focus on key features of the inputs, look ahead along short trajectories, and place tripwires to trigger certain behaviors. 

This approach proposes a model addressing  the interpretable box design problem for DRL methods applied in visual inputs, where the policy is modeled by an LSTM network. The method focuses on providing global policy explanations. The model exploits LSTM states to produce queries, as well as keys and values tensors produced by the visual input: Thus, we could say that the query network and the vision core serve as the $f(\cdot)$ to feed the interpretation component. The interpretable model $c_g$ is a soft attention model and  the explanation logic $\epsilon_g$ is implemented by visualizing the attention maps, showing the original input
frames and super-imposing the attention map for each attention head. In addition, saliency maps, visualizing the relative dominance of ``what" and ``where" parts of the query are being used, to inspect where the agent focuses attention.

%Up to now: attention mechanisms only for visual agents

\subsubsection{Conservative Q-Improvement}
\label{sec:4.2.8-ConservativeQImprovement}

Decision trees that represent policies span the state space and may be larger than required, even if they do express the action value function quite accurately. A. M. Roth et al. in \cite{ConservativeQImprovement} aim to provide an answer to the question ``what should be the size of such a decision tree" to make it accurate and interpretable? This work is based on the hypothesis that the size should be that, which if increased further it does not increase significantly the estimated discounted future reward of the overall policy.

In doing so, Roth et al. in \cite{ConservativeQImprovement} propose the Conservative Q Improvement
(CQI) RL algorithm. CQI learns a policy
in the form of a decision tree, in any domain with discrete actions and multidimensional state space. In contrast to other RL methods (e.g., \cite{Pyeatt}) that learn a decision tree policy representation, this approach uses a lookahead approach that predicts
which split will produce the largest increase in reward.
The method is strictly additive: Initialized with a single leaf node, over
time, it creates branches and leaf nodes by replacing existing leaf
nodes  with a branch node and two child leaf nodes. A branch node represents abstract states and indicates actions to be taken using the $Q$ function. A leaf node is split (and turned into a branch node) only when the expected discounted future
reward of the new policy would increase above a dynamic
threshold, moderated by the node visit frequency and the dynamic split threshold.

CQI is evaluated in the RobotNav domain, where a robot must navigate in a 2D environment to reach a goal location, avoiding obstacles. It is shown that the method is able to produce substantially smaller trees compared to the method of \cite{Pyeatt}, with smaller variance and of greater reward.  No explanation logic $\epsilon$ is proposed in reference \cite{ConservativeQImprovement}, and this  article does not evaluate the method in terms of interpretability or explainability.

CQI follows the interpretable box design paradigm: It aims to learn an interpretable policy model $c_g$ in a direct way towards increasing explainability. 

\subsubsection{Hierarchical interpretable reinforcement learning}
\label{sec:4.2.9-HierarchicalinterpretableRL}

T. Shu et al. in \cite{HierarchicalinterpretableRL} propose a framework for efficient
multi-task RL, training agents to employ hierarchical
policies: These policies support an agent to decide when to use a previously learned policy and when to
learn a new skill. Each learned task has a corresponding human language description, thus represented in an interpretable way. Through the hierarchical model proposed, one
may i) accumulate tasks progressively from a terminal policy to a top-level policy, and ii) unfold the
global policy from top-level to basic actions, using human language descriptions. This approach not only learns the language grounding for visual knowledge and policies,
but is also trained to utter human instructions as an explicit explanation of its decisions to
humans.
%In order to better track temporal relationships between tasks, a stochastic temporal grammar (STG) model is trained on the sequence of policy selections (previously learned skill or new skill) for positive episodes. Integrating the STG into the hierarchical policy boosts efficiency and  accuracy by explicitly modeling commonsense world knowledge.

More specifically, training progresses at stages, increasing the set of tasks from $\mathcal{G}_0$ to $\mathcal{G}_k$ at stage $k$, and the learned policies from $\pi_0$ for $\mathcal{G}_0$, to $\pi_k$ for $\mathcal{G}_k$. At stage $k$, the  policy $\pi_k$
is defined by a hierarchical policy that consists of four sub-policies: a base policy for
executing previously learned tasks (the policy at stage $k-1$), an instruction policy that manages communication between the
global policy and the base policy (mapping a task $g \in \mathcal{G}_{k}$ to a base task  $g' \in \mathcal{G}_{k-1}$), an augmented flat policy which allows the global policy to directly
execute actions instead of using a base task, and a switch policy that decides whether the global policy will primarily rely on the
base policy or the augmented flat policy.
A stochastic temporal grammar (STG) model is trained on the sequence of policy selections (previously learned skill or new skill) and focuses on priorities among tasks: It interacts with the hierarchical policy through modified
switch policy and instruction policy. 
%This amounts to treating the past history of switches and instructions in positive episodes as a guidance on whether the hierarchical policy should defer to the base policy to execute a specific base task or employ its own augmented flat policy to take a primitive action. 
Learning at each stage is a 2-phase curriculum learning task: Policies in each phase are learned with an advantage actor-critic (A2C) off-policy method, assuming that the terminal policy $\pi_0$ has been trained, with base skill acquisition  (learning to use previously learned skills)  and novel skill acquisition phases (learning to rely on the augmented flat policy for executing novel tasks).

The approach is evaluated  in a two room environment in Minecraft with 24 tasks in total. Experiments aim mainly to show learning efficiency and generalization abilities. Hierarchical plans of several tasks generated by global policies learned by the
full model are visualized. However, these visualizations are mainly used to show agent's learning efficiency, rather than the provision of explanations.

Concluding, this approach follows  the interpretable model design paradigm, training  interpretable models that are hierarchical plans, where skills and tasks are described in natural language based on human instructions.

%%%%%%%%%%%%%%%%%%%%%%%%%%%%
%\subsubsection{Mimicking}
%%%%%%%%%%%%%%%%%%%%%%%%%%%%

\subsubsection{LVIN: Imitation learning using the Value Iteration Networks approach}
\label{sec:4.2.10-LVIN}

The Logic-based Value Iteration
Networks (LVIN) proposed by T. Leech in \cite{LVIN} combines imitation learning and learns to represent
problems as compact finite state automata (FSA) with human-interpretable logic
states. The aim is to facilitate understanding and manipulation of
the learned model, towards adding trust and flexibility to robotics applications trained with
LVIN.

LVIN produces a policy which describes desired actions in terms
of transitions between human-interpretable logic states. In doing so, the planning
problem posed by the decision layer is decomposed into two pieces, a high level finite state
automaton (FSA) that corresponds to high level sub-goals the robot follows (described
in logic) and a low level Markov Decision Process (MDP) describing the motion of the
robot in the physical environment. The FSA learns the rules governing the robot’s
behavior in terms of interpretable and manipulable logic states, and transitions, called propositions. 
LVIN 
%factors the overall system MDP into an FSA to describe the high level logical process and a low level MDP describing the physical environment, as in Value Iteration Networks. The addition of high level logic states to the description of the world ends up with a total system MDP described by the tuple $(S \times F, A \times L, R, T, \gamma)$, where $(S,A,R,T,\gamma)$, with $S$ the state space, $A$ the action space, $R$ the reward function, $T$ the transition function and $\gamma$ the discount factor, is the MDP that the Value Iteration Network approximates from expert demonstrations, and $(F,L,\emptyset,TM, 1)$ is the factored high level FSA MDP, given the logic states $F$, propositions $L$, and transitions at the logic level $TM$. LVIN 
comprises separate Value Iteration Networks for each FSA state, where learning the state transitions  associated with the high level FSA involves encoding the order of propositions to trigger given the current FSA state, and learning the $P(s'|s,a)$, $s \in S$ and $a \in A$, associated with the low level MDP. LVIN learns the entire model via an iterative update process, similar to Value Iteration Networks, where the update procedure is encoded in a deep NN.

Evaluation aims to show the success rate (i.e., the percentage of correctly completed tasks) of LVIN compared to a CNN trained with a map as input and the desired action as output. The method has been evaluated in a lunchbox packing and in a cabinet checking problem, with no emphasis on evaluating interpretability or explainability.

LVIN addresses the model explanation problem for discrete actions and in a 2D space. The interpretable global model $c_g$ is the high level FSA. In this case, the function  $f(\cdot)$ provides the ``raw" model (deep NN) parameters, i.e., making no interpretation. Since the high level FSA and the low level MDP are trained "seamlessly" taking as input the same expert demonstrations, learning the interpretable model can be considered as a mimicking process.
LVIN does not propose any specific explanation logic for realizing the function $\epsilon_g$. Abstract policies in terms of logical states and transitions explain low level policies, assuming that these abstractions are interpretable by humans. As authors point out ``the current LVIN framework ... is still somewhat limited by the fact that it requires a hard coded
logic oracle for training".

\subsubsection{Interpretable DRL with Linear Model U-Trees}
\label{sec:4.2.11-interpretableDRLwithLMUTs}

G. Liu et al. in \cite{interpretableDRLwithLMUTs} introduce Linear Model U-trees (LMUTs) to approximate
NN predictions for DRL agents. 
An LMUT is learned by an on-line
algorithm that is well-suited for an active play setting, where the mimic
learner observes an ongoing interaction between the DRL and the
environment. The article shows how the interpretable tree structure of an LMUT facilitates analyzing feature influence, extracting rules, and
highlighting the super-pixels in image inputs.

U-tree \cite{U-tree} \cite{CUT}  is a classic online RL method which represents a $Q$ function using a tree structure. 
%A U-tree takes a set of observed feature/action values as input and maps it to a state value (or to a Q-value). 
In addition to that, continuous U-tree (CUT) 
\cite{CUT} dynamically generate
a tree-based discretization of the input signal and estimates state transition
probabilities by retaining transitions in every leaf node.  
To strengthen the generalization ability of these representations, as well as their learning efficiency,
Linear Model U-Trees (LMUT)  \cite{interpretableDRLwithLMUTs} add a linear model to each leaf node. 
This can be trained either in a batch  mode or in an active play setting. LMUT applies Stochastic Gradient Descent to update the linear models, given some memory of recent input data stored on each
leaf node.  In the active play setting the method exploits transitions in the form of $(state, action, Q-value, next-state, reward)$, where the $action$ is decided by the DRL model, together with the ``soft" $Q-value$, while the $reward$ is provided by the environment.

LMUT builds an MDP from the interaction data between the environment and
a deep model. Compared to a linear Q-function approximator, a Linear Model U-Tree (corresponding to an action) defines
an ensemble of linear Q-function models, one per leaf node and one for each state partition cell. Since each
Q-value prediction comes from a single linear model, the prediction can be
explained by the feature weights of the model.

This approach, has been evaluated towards mimicking a DQN model, and compared against CART, M5 with regression tree, Fast Incremental Model Tree (FIMT) baseline and with Adaptive Filters (FIMT-AF). The evaluation environments include Mountain Car, Cart Pole and Flappy Birds all with discrete actions but with discrete or continuous states.
Evaluation concerns approximating the Q values in DQN, reporting on the number of the tree leaves constructed. Furthermore, the sampling efficiency of LMUT is evaluated by computing the correlation and testing error of LMUT, as more transitions are provided. Interestingly, LMUT models are evaluated in terms of the Average Reward Per Episodes by performing the tasks (e.g., playing games) in a direct manner: It is reported that LMUT among other mimic models achieves performance which is closest to the DQN. As far as interpretability is concerned, authors propose (a) a way to measure splitting features' "importance" using the total variance reduction of the $Q$ values, (b) a way to extract and present rules in the form of partition cells, each cell representing a range in specific splitting features, and (c) a way of highlighting pixels (when DRL takes  raw pixels as input) while explaining local decisions. However, these options are not evaluated in any setting with human subjects.

Concluding, this is  a mimicking approach, where the interpretable model is trained with  transitions caused by the interactions of the DRL agent with the environment, also consulting DRL model decisions and values, via the function $f(\cdot)$. The interpretable model represents the agent policy, and as mentioned above, it can be exploited in several ways to provide interpretations. The article does not suggest any particular function $\epsilon$ implementing an explanation logic.

\subsubsection{Policy level explanations for Reinforcement Learning}
\label{sec:4.2.12-PolicyLevelExplanations}

To address the need of explaining sequences of decisions, N. Topin and M. Veloso in \cite{PolicyLevelExplanations} introduce Abstracted Policy Graphs (APGs).  APGs  are Markov
chains of abstract states and are assumed interpretable representations of policies, concisely summarizing them, so that individual decisions can be explained
in the context of expected future transitions. The authors propose the APG Gen method for generating APGs
for deterministic policies given a learned value function and
a set of observed (potentially off-policy) transitions. % There are no restrictions on the value function learned.  

An APG is a Markov chain over abstract states where edges are transitions induced by actions from the original MDP. The basic assumption here is that, if the agent’s transitions
between grounded states are approximated using a Markov
chain between abstract states, then grounded states  which are
\textit{treated similarly} (thus, can be \textit{treated interchangeably}) are readily identified (i.e., in groups) and the agent’s transitions
between abstract states can be predicted. Interchangeable ground states grouped into abstract states should have similar future outcomes in terms of rewards, which are assumed known, and the importance of all features for a set of similar states should be low. The  importance of a feature is calculated by the Feature Importance Ranking Measure (FIRM), as   the variance of the conditional expected score of a state $s$ for that feature, with respect to
an arbitrary function $g(s)$. This score is the average value of
$g(s)$ for all states, where that feature takes a specific value $v$.

APG Gen \cite{PolicyLevelExplanations} uses states' similarity based on actions taken under the policy, to repeatedly divide abstract
states along important features, and then  computes transition probabilities between them. Important features can be trivially recorded 
for each abstract state, allowing for humans to inspect which features affect agent's decisions, also providing a summary of an abstract state and of the corresponding ground states.

APG is evaluated in an abstraction of a production task for a multi-component item using a number of manufacturing steps, in deterministic and stochastic settings. Policies are computed using value iteration. In providing local explanations, APG is evaluated in terms of the features' prediction accuracy (important vs. not important) as a function of the portion of states sampled: This provides a view on how APG generalizes on features' importance, beyond the states seen. Additionally, 
n-hop action predictions are evaluated to measure the error caused by assuming that only single actions are performed for a transition between abstract states. This is measured by the action prediction agreement (predicted vs actual) for increasing time horizon.
Finally, the size of ``explanations" are measured by the number of abstract states (nodes in APG) against the number of grounded states in the MDP (reported sub-linear).
No evaluation with human subjects in simulated or real-world settings are reported.

Concluding, APG Gen  is a mimicking approach,  with the objective of providing an interpretable model of a learned policy, considering a trained policy model. That model is realized by an APG.  APG Gen  takes as input transition  samples  $(s,a,s')$ from the learned policy, the value function, as well as features' importance.  APG is the interpretable model, and local explanations may be provided consisting of the features that are important for a specific state. Reference \cite{PolicyLevelExplanations} does not propose an explanation function $\epsilon$ for providing interpretations' surface representations.

\subsubsection{PIRL: Programatically interpretable reinforcement learning}
\label{sec:4.2.13-PIRL}

PIRL (Programmatically Interpretable Reinforcement Learning) is an RL framework \cite{PIRL} designed to generate
agent policies that are represented using
a high-level, domain-specific programming
language. Such programmatic policies are not as performant as those from DRL methods, but are interpretable and being amenable to verification
by symbolic methods. A. Verma et al. in  \cite{PIRL} propose the Neurally Directed Program
Search (NDPS)  method for solving the challenging non-smooth
optimization problem of finding a programmatic
policy with maximal reward, minimizing the distance from an  “oracle”  DRL policy.

Specifically, considering a Partially Observable
Markov Decision Process (POMDP) setting, a functional language, and given a policy sketch  that syntactically defines a set of
programmatic policies, the objective is to
find a program in this set with maximal long-term reward. 
To restrict the search, the NDPS algorithm performs directed policy search guided by an oracle policy learned by a DRL method, via local search. NDPS measures the ``distance" between the outputs of a programmatic policy and the outputs of the oracle policy, given a set of ``interesting" histories which are enriched with additional histories generated by the programmatic policy (a technique that in \cite{PIRL} is called ``input augmentation"). Further optimization improves the programmatic policies found (e.g., being shorter).

PIRL is evaluated on the task of learning to drive
a simulated car in the TORCS car-racing environment, using as oracle a policy computed by the Deep Deterministic Policy Gradient (DDPG) algorithm: NDPS policies are compared to human-readable policies, also showing that
PIRL policies can have smoother trajectories, and
 better generalization abilities compared to the
policies discovered by DRL. No evaluation of the explainability or interpretability of PIRL is provided, with or without human subjects.

Concluding, PIRL with NDPS is a mimicking process, exploiting agents' interactions with the environment to find optimal interpretable programmatic policies that are presented as functional programs. 

\subsubsection{Understanding Decisions by Interpretable Policy Learning}
\label{sec:4.2.14-ExplainingByImitation} 

INTERPOLE is an imitation learning method that jointly models the agent's policy (belief-action mapping)  and belief-update process, towards interpreting demonstrated behaviour. As such, it can be used to model any ``expert" demonstrated policy, without assuming unbiasedness of agent's beliefs or optimality of policies. In so doing, A. H{\"u}y{\"u}k et al. in \cite{ExplainingByImitation} propose an inherently interpretable imitation learning method, locating factors that contribute to individual decisions, in a language that is readily understood by domain experts, accommodating partial observability and operating offline.

INTERPOLE aggregates sequential observations through agent's subjective belief-update process (decision dynamics) and probabilistic belief-action mapping for determining actions (decision boundaries). The objective is to model the most likely parameterizations $\theta = (T,O,b_1,\eta,\{\mu_a\}_{a \in A})$, drawn from some prior $\mathbb{P}(\theta)$, i.e. the likelihood of $\theta$ with respect to a given set of demonstrated observation trajectories $\mathcal{\bar D}$ and unobserved state trajectories $\mathcal{D}$:  $T(s_{t+1}|s_t, a_t)$ is the state transition probability, $O(z_t|s_t, a_t, s_{t+1})$ is observation's probability after a transition to a new state, $b_1(s)$ is the probability of being in state $s$ in $t=1$, the smoothness of transitions between decision boundaries is tuned by the temperature factor $\eta$, and $\{\mu_a\}_{a \in A}$ are actions' mean vectors inducing decision boundaries over the belief space. 
Parameters $\theta$ are jointly determined by means of an expectation-minimization (EM)-like algorithm for maximizing $\mathbb{P}(\mathcal{D})$, given that $\mathcal{D}$ is not known. Estimating jointly the parameters $\theta$, INTERPOLE learns agent's decision and belief dynamics (not the environment dynamics) offering the most plausible explanation of how the agent reasons.

INTERPOLE has been evaluated using simulated and real world demonstrated trajectories for disease diagnosis, focusing on interpretability, accuracy of learned policies, and subjectivity towards recovering interpretations of behaviour even if the agent is biased. For interpretability, INTERPOLE has been evaluated by nine clinicians focusing on the utility of representing possibly subjective action-belief trajectories instead of raw action-observation trajectories, and on understanding policies by means of suboptimal decision boundaries, instead of using a reward function: The majority of clinicians  preferred the explanations offered by INTERPOLE.
Regarding accuracy, INTERPOLE is evaluated in modelling the belief-update process and the belief-action mapping (policy), learning high quality models driving agent's subjective reasoning.

INTERPOLE \cite{ExplainingByImitation} addresses the policy explanation problem, offering a method for imitating  agents' possibly suboptimal policies without assuming unbiasedness beliefs and objective reasoning. The interpretable model $c_g$ comprises decision dynamics and decision boundaries and is specified by $\theta = (T,O,b_1,\eta,\{\mu_a\}_{a \in A})$ without assuming a fully observable setting, given a set of demonstrated action-observation trajectories $\mathcal{\bar D}$. The explanation logic $\epsilon$ visualizes a belief simplex in low dimensional spaces, showing decision boundaries and belief trajectories.   

\subsubsection{Automatic discovery of interpretable planning strategies}
\label{sec:4.2.15-InterpretablePlanningStrategies} 

With the aim to assist human experts in decision making, J. Skirzy{\'n}ski et al. in \cite{InterpretablePlanningStrategies}  introduce the Adaptive Imitation-Interpretation (AI-Interpret) algorithm for transforming policies into sets of interpretable descriptions by means of decision rules  that have a  disjunctive normal form (logical program policies), presented as flawcharts.

Specifically, given a set of demonstrated trajectories $\mathcal{D}$ produced by the original policy, a domain specific language specifying the predicates for constructing the logical program policies, a parameter $\alpha$ defining the threshold  at which the discovered policy (represented by a logical formula)  is at least better than the original policy, and the maximum rule length ($d$) of the logical formula,  the objective is to find the logical formula that maximizes the number of demonstrations in $\mathcal{D}$ that it can imitate w.r.t. $\alpha$ and $d$.
AI-Interpret uses the predicates to separate the set of demonstrations into clusters. Clusters (whose number is determined automatically) are evaluated by means of a heuristic value that shows the similarity of trajectories in the cluster and how representative the cluster is. Doing so, the algorithm can consider increasingly smaller sets of demonstrations (disregarding less important clusters) and employ Logical Program Policies (LPP) \cite{LPP} in a structured search searching for the simplest logical formula w.r.t. $\alpha$ and $d$  that imitates most of the demonstrated trajectories. Finally, the formulae are transformed into decision trees, and then visualized as flowcharts that people can follow to execute the strategy.

AI-Interpret is evaluated in three types of challenging planning tasks.  The  central question is whether discovered flawcharts can support
decision-makers in the process of planning. To achieve that, authors in  \cite{InterpretablePlanningStrategies} perform one large behavioral experiment for each of the three types of tasks and a fourth experiment in which they evaluate whether AI-Interpret improves human decision-making against conventional training (i.e., by providing feedback). Results show that the proposed approach (a) allows people to largely understand the automatically discovered strategies and  use  them  to  make  better  decisions, and (b)  is more effective than conventional training, improving human decision-making.

Overall, AI-Interpret \cite{InterpretablePlanningStrategies} addresses the policy explanation problem, discovering policies described by means of logical programs. This is a mimicking approach, as the input it requires are demonstrations from the closed-box RL policy model. The interpretable model $c_g$ is the LPP discovered, and the explanation logic $e_g$ transforms LPP to flowcharts.

\subsection{Solving the objectives explanation problem}

\subsubsection{Enabling robots to communicate their objectives}
\label{sec:4.3.1-CommunicateObjectives}

S. H. Huang et al. in \cite{CommunicateObjectives} conjecture that, end-users, to understand and predict what the agent will do, need to
have an accurate mental model of the agent's objective function. To reduce the time needed for this otherwise time-lengthy process, authors propose to model how people infer objectives from observed behavior, in
order agents to demonstrate behaviors that are maximally informative. 
%Authors in \cite{CommunicateObjectives} introduce factors to define candidate models of human inference,
%, and show that certain models have the capacities to prescribe the provision of exemplary robot behaviors that better enable users to anticipate what the robot will do in novel situations. However, it is shown that choosing the appropriate model is a key aspect, and candidate models do not fully capture how humans extrapolate from examples of robot behavior. These findings are leveraged  to propose a stronger model of human learning, and conclude by analyzing the impact of different ways in which the assumed model of human learning may be incorrect.

Following an  algorithmic teaching approach, authors in \cite{CommunicateObjectives} define candidate models of human inference regarding robot’s objective
function, given robot's optimal behavior in specific environments: These demonstrations increase the probability
of humans to infer the correct objective function. The objective function is assumed to be a linear combination of (possibly complex functions on) environment features  in $M$, weighted by  parameters $\phi^*$. Features are supposed to be known by humans (although not always the case), who need to learn the true reward parameters. This involves searching for a sequence of environments $E_{1:n}$, such that when the person observes
the optimal trajectories in these environments, their updated
belief places maximum probability on the correct reward parameters.
The challenges tackled to solve this optimization problem  are as follows:  (a) Humans are likely to be approximate in their inference,  while (b) the agent needs to incorporate a model of this approximate inference, (c) encouraging full coverage of all possible strategies that the agent is capable of adopting.
Updates of humans' beliefs for the reward  parameters can be modelled either via an inverse reinforcement learning technique (exact method), %eliminating all objective functions that do not assign maximum reward to observed behaviour, 
or via Bayesian inference (approximate method), modelling how probable humans consider trajectories to be, given their beliefs on the reward function.  To address the coverage challenge, a new term is added to the formulation of the optimization problem, ensuring that no extra trajectory examples are selected unless these examples increase significantly the probability of inferring the true reward parameters. Extra examples are chosen to be the best with respect to the best approximate model.

Experimental results evaluate how approximate and exact inference models perform in a driving simulation environment. The aim is the system to provide to participants examples of how the car drives, with the goal of being able to anticipate how it will drive when they ride in it. The analysis of models is done with ideal users and with human subjects. Analysis shows that while not just any approximate-inference model is better than the exact one, choosing a suitable approximate model helps significantly. 
 The user study on coverage concludes that coverage with the right model of approximate inference have a significant advantage over RL models assuming exact inference users.

Concluding, reference \cite{CommunicateObjectives} provides methods that address the objective model explanation problem, focusing on teaching the agent's reward function true parameters via examples. The proposed approach explicitly incorporates a model of the human beliefs on reward parameters and aims to provide a view on the true objective function. % and is implemented by the example selection algorithm, in conjunction  to the update of human beliefs over reward parameters. 
While the interpretable model can be considered to be the linear objective function, no explanation logic is provided here.

\subsubsection{Explaining agent behaviour through intended outcome}
\label{sec:4.3.2-IntendedOutcome}

H. Yau, C. Russell and S. Hadfield in \cite{WhatDidYouThink}, aim to interpret decisions  by describing agents' intended outcome. This method applies to RL methods that estimate values off/on policy, given discrete states and actions and a deterministic MDP. The approach is demonstrated to off-policy $Q$-learning and aims to recover the future events that impact the $Q$-values learned.  

This approach learns a {\em belief map}  of discounted  expected sum of state visitation \textbf{H}. \textbf{H} is updated when a Q-value is updated, as follows: 
$\textbf{H}(s_t, a_t) = \textbf{H}(s_t, a_t) +\alpha(1_{s_t,a_t} + \gamma\textbf{H}(s_{t+1}, argmax_aQ(s_{t+1}, a_{t+1}))-\textbf{H}(s_t, a_t))$, 
 where $1_{s_t,a_t}$ denotes an indicator function with 1 if the agent is at $(s_t,a_t)$ and 0 otherwise.

Given a deterministic reward function $R_{s_t,a_t}$, the belief map \textbf{H} is consistent with $Q$ if  $vec(\textbf{H}(s,a))^Tvec(R(s,a))=Q(s,a), \forall (s, a) \in S \times A $. Authors prove that if \textbf{H} and $Q$ are zero-initialized, then \textbf{H} and $Q$ are consistent for all iterations of the algorithm in the tabular case. For deep Q-learning methods, there is no guarantee for  consistency of \textbf{H}  with Q.

The proposed approach is evaluated in  domains with deterministic or stochastic reward functions and with continuous or discrete state-action space. In environments with stochastic rewards and continuous states, new discrete states are added, or states are discretized. The method is demonstrated to work effectively for tabular and deep Q-learning agents. 

This is a generic approach that addresses the objective model explanation problem, providing projections of future intended trajectories of agents.  The interpretable model is provided by the belief maps, and explanations are provided by means of belief maps visualizations showing the value of each intended future state-action pair. The method is limited to low-dimensional state-action spaces.

\subsubsection{Distal Explanations for explainable reinforcement learning}
\label{sec:4.3.3-DistalExplanations}

P. Madumal et al. in \cite{DistalExplanations} emphasize on the role of opportunity chains for explaining agents' behaviour, and 
introduce a distal explanation model that can
generate opportunity chains as explanations for model-free RL
agents. An opportunity chain takes the form of ``A enables B and B causes C", where B is the distal event or action, and can inform the explainee about long term dependencies
between events, when certain events enable others.

Reference \cite{DistalExplanations} proposes a distal explanation model that
learns opportunity chains using a recurrent neural network (RNN). As distal
explanations by themselves would not make a complete explanation, this approach uses action influence models \cite{CausalLens} that approximate the causal model of the environment relative to actions taken by an agent, to get the agent’s
goals. It further improves upon action influence models (specifying the effects of actions in features) using
decision trees instead of structural equations used in \cite{CausalLens} (presented in Section \ref{sec:4.4.7-CausalLens})  to represent the agent’s policy.
The decision tree policy model is trained concurrently with the RL policy model, assuming a model-free algorithm and exploiting state-action samples using an experience replay buffer. The tree is constrained to have a max number of leaves, equal to the agent actions.
The same dataset of samples exploited for learning the decision tree model is exploited by a prediction RNN to approximate distal actions and their cumulative rewards. 

The accuracy of distal prediction and counterfactuals is evaluated in six benchmark domains using different
model-free RL algorithms. Benchmarks have a mix of complexity levels
and causal graph sizes (number of actions and state
variables). Results show that the distal explanation model is
robust and accurate across different environments and algorithms.
Furthermore, experiments with humans using RL agents report on  task prediction (i.e., understanding of the agent) and subjective explanation
satisfaction. Agents are trained to solve  1)
an adversarial task; 2) a search and rescue task; and 3) a human-AI
collaborative build task; all in Starcraft II. The results obtained for explanation quality show that distal explanations perform
substantially better than baselines in human-agent collaborative tasks.

Concluding, this is a mimicking process, where the interpretable models (decision tree and opportunity chains) are trained by exploiting RL agent's interactions with the environment. The interpretable models comprise a  policy model $c_g$ using a decision tree, and a causal model representing the objectives ($c_o$), as well as  the importance of specific responses using opportunity chains. The explanation  logic $\epsilon$ exploits models to produce explanations in  natural language.

\subsection{Solving the outcome explanation problem}

\subsubsection{Reward Decomposition}
\label{sec:4.4.1-RewardDecomposition}

The approach proposed by Z. Juozapaitis et al. in \cite{RewardDecomposition} decomposes rewards into
sums of semantically meaningful reward types,
so that actions can be compared in terms of
trade-offs among the types, with a focus on explainability. In particular, this work
introduces the concept of minimum sufficient explanations for compactly explaining why one
action is preferred over another in terms of the reward types. 

Specifically, assuming that the MDP formulation incorporates a set of reward components/types
$C$ and defining a vector-valued reward function with a component for each $c \in C$, $R_c(s, a)$,  the agent objective is to optimize the overall (mixed) reward function $R(s, a) = \sum_{c \in C}R_c(s, a)$.
This vector-valued reward allows for defining
a vector-valued $Q$-function, with action-value components  $Q_c(s, a)$ for rewards of type $c$. The overall Q-function is defined to be $Q(s, a) = \sum_{c \in C}Q_c(s, a)$. This results in a straight-forward heuristic tabular algorithm updating $Q$ values in a standard way for each of the components, called HRA. The definitions are extended to DRL settings, where each $Q_c$ is parameterized by means of $\theta_c$. This results into DRL algorithms for decomposed reward (dr), drDQN and drD-SARSA.

To gain insight
into why an agent prefers action $a_1$ over $a_2$ in state $s$,
i.e., $Q(s, a_1) > Q(s, a_2)$, this  approach uses the difference
explanation (RDX) as the difference of the decomposed
$Q$-vectors.  RDX is exploited to find the minimal sufficient explanations $MSX^+$ and $MSX^-$ for each pair of actions and state, indicating the ``critical" positive and negative reasons (in terms of reward types), respectively for the preference.

Case studies in two environments
 illustrate the potential of visual explanations in the
hands of an RL practitioner. Authors show how decomposed reward methods support spotting
certain ``bugs" in the agent's action values and even help
identify a more subtle issue related to the interaction of the gradient optimizer and the DRL loop.

Concluding, this approach addresses the outcome explanation problem, focusing explaning decisions on actions against other actions locally, i.e., in specific states, exploiting reward types. The interpretable model $c_l$ takes the form of tuples $(MSX^+, MSX^-)$ per state and applicable action couples, exploiting  (decomposed) action values provided by $f(\cdot)$. The explanation logic $\epsilon_l$ is implemented by means of bar charts on features.

The reward decomposition approach has been proposed as the basis for a set of decision-making tasks  in \cite{StrategicTasks}, in environments that naturally provide multiple reward signals: The goal is to enhance explainability by providing high-level abstractions for sequential tasks.

\subsubsection{Autonomous Policy Explanation}
\label{sec:4.4.2-AutonomousPolicyExplanation}

B. Hayes and J. A. Shah in \cite{AutonomousPolicyExplanation} aim to enable an autonomous agent to reason
over and answer questions about its underlying control logic $L$,
independent of its internal representation. First, it learns  a domain model of the agent’s
operating environment from real or simulated demonstrations, exploiting programmer specified code annotations for actions, and variables (state parameters/attributes) affected by actions.
Within this learned domain model, this approach uses statistics computed over data extracted from continued observations 
%of the trained agent’s software execution traces 
to construct a behavioral model that approximates the agent’s control logic. The article shows applicability of the proposed approach in tabular Q-learning, deep Q-learning  and hand-coded controller logic.   
%A Boolean algebra over the space of defined planning predicates is used as a basis for  grounding state regions in natural language, succinctly.

To explain the policy $L$ given a query $q$, this work provides an interpretation function $c$ to provide a natural language response $Response$. $L$ is constrained to state parameters and annotated functions. $Response$ is constrained to a template-based approach using natural language.

The interpretation function $c$ is defined to  be
$c = Summarize\_attributes \textit{ o }Resolve\_states \textit{ o }Identify\_question$,
\noindent where, $Resolve\_state$ resolves the question identified by $Identify\_question$ to a set of problem states (context), and $Summarize\_attributes$ summarizes attributes of problem states into a comprehensive representation. The $Resolve\_state$ realizes the function $f(\cdot)$, providing a representation of relevant parameters to resolve a question. Depending on the question type (see below), this may take the form of searching and gathering  state parameters' from states in a region, or from similar states given a specific state, or from near-by states.   Questions concern when an agent will behave in a specific way, how it will behave under specific conditions, and deviations from the expected behavior.   Question-answering is template driven using a response resolution algorithm per query type. Depending on the question type, $c$ provides an explanation for a specific response (i.e., locally) at the granularity of policy actions, explaining the control logic in specific contexts (i.e., sets of states). 
The  explanation logic $\epsilon$ is implemented by the $Compose\_summary$ function and is realized by means of Boolean classifiers using communicable predicates, which are similar to STRIPS-style planning predicates, associated with natural language descriptions.

%However, the interpretation has not been shown to provide explanations for  trajectories produced by the policy (i.e., for responses at larger granularity and scale), nor for the objectives that the agent aims to achieve.

This work allows human collaborators
to shape their expectations on agents' behaviour without requiring a full understanding
of an agent’s logic, and facilitates the debugging
of aberrant behaviors by explaining differences
between the conditions under which particular actions occur.
The article demonstrates the applicability of the proposed method within three
representative robotics domains with discrete and continuous state parameters, in single/multi agent settings, using different types of controllers. However, experiments focus on the applicability of the method, rather than on the accuracy/fidelity and conciseness of the explanations provided in any specific  human-robot collaborative setting.

\subsubsection{Transparency Communication for Reinforcement Learning}
\label{sec:4.4.3-TransparencyCommunicationforRL}

N. Wang et al. in \cite{TransparencyCommunicationforRL}, as a continuation of previous work  on static models \cite{TransparencyCommunicationforRL}, discuss the design of model-based and model-free RL for robots in a human-robot simulation testbed. The goal is to  provide communications for the robot's decision making update process towards interpretability and repairing trust. This is achieved by interpreting components of the robot's decision making and learning process, assuming that the robot bases its decisions on a POMDP model. %$(S,A,P,\Omega,O,R)$, where $S,A,P,R$ are as in MPD, $\Omega$ is a set of possible observations and $O$ is the observation function that - given a state-  provides the probability for the robot to get any observation.
%Such an interpretation is provided via natural language static templates providing information about (a) agent's current beliefs about the state of the world (maybe using also probabilities to represent uncertainties), (b) specific actions decided, (c) relative likelihood of possible responses based on the transition model, (d) sensing abilities via communicating specific observations, (e) the observation model, and finally, (f) the expected reward given agent's current action, relying on factored rewards.

Considering a dynamic POMDP
and assuming a model-based RL that updates the POMDP model, the proposed method  interprets a static POMDP after each update, informing the teammate about changes. 
This work focuses on updating the observations model, i.e., the probability for the robot to get any observation. Additionally, assuming a model-free RL, % that updates $Q$ values based on robot's experiences
 $Q$ values account for changes in a possible model, but with no explicit representation of these changes and model.
To address this last issue, authors propose finding a potential POMDP that is consistent with the policy arrived at by the model-free RL. To do that, they compute the optimal policy for that POMDP exploiting the assumption of a piecewise-linear model \cite{PieceWiseLinearFunctions}, representing that policy by means of a decision tree, which is compared to the RL policy. If it matches, this POMDP is used to provide explanations.

This article does not report on  the efficacy of explanations with human-subjects. As authors also point out, this must be done with different permutations of explanations' aspects, also providing insight into what aspects should
be included in the explanations and how to present them.

Concluding, this approach provides interpretations of POMDP components, with no access to the learning process itself. We can consider that functions $f(\cdot)$ provide updates on agents' beliefs, transition models, or updates in the $Q$ values, while the interpretation process updates the existing POMDP, or searches for a potential matching POMDP, which serves as the interpretable model. Function  $\epsilon$ provides explanations on POMDP components' updates using natural language static templates. In addition to revealing information about individual POMDP components,  templates can be created informing about the overall model and how the learning arrived at that model, also leveraging the interpretable policy models generated by exploiting the piecewise-linear assumption.

\subsubsection{Transparency \& Explanation in DRL}
\label{sec:4.4.4-TransparencyandExplanationInDRL}

R. Iyer et al. in \cite{TransparencyandExplanationInDRL} report on the interpretability of DRL Networks (DRLN) in visual tasks, proposing the Object-sensitive
DRL (O-DRL) method that (a) incorporates explicit object recognition
processing into DRL models, (b)
forms the basis for the development of object saliency maps,
%to visualize of internal states of DRLNs, 
and (c) can be incorporated
in any existing DRL framework such as DQN and A3C.
The goal here is to produce intelligible visualizations of agents' state and behaviour, focusing on objects saliency, i.e., on the influence of objects features in agents' decisions.

The  proposed  O-DRL  approach uses a deep neural network (CNN) that takes as input object channels and original images to predict $Q$-values. Object channels represent types of objects recognized in an image and incorporate objects' features. To compute object saliency, this  proposal masks the object and computes the $Q$ values on the new (perturbed) state, and the difference of new values to the $Q$ values in the original state.

Authors report on experiments with humans, regarding the behaviour of a Pacman agent. The goals
of the experiments were to (a) test whether object saliency
maps contain enough information to allow humans to match
them with corresponding game scenarios, (b) test whether
participants could use object saliency maps to generate reasonable
explanations of the behavior of the Pacman and
(c) test whether object saliency maps allow participants to
correctly predict the agent’s next action. 
Experiments include (a) a matching task, where participants match saliency maps to agents' behaviour;
%, providing also a teleological explanation explaining why the agent acted as it did;
and (b) a movement prediction task from video clips with or without saliency maps. %Participants also asked to give an explanation for their prediction which includes their judgment as to which elements of the game influenced the agent's decision.
Results show that  object saliency maps can be linked to corresponding
game scenarios by participants (thus, they do provide a basis for visual explanations) but, as results from the prediction task suggest, the exact use of salient maps must be further investigated in conjunction to providing rich contextual information about each situation.

This is an approach that addresses the outcome explanation problem in visual tasks, focusing on the influence of objects in agents' decisions in specific states. The proposed interpretation method is realized by the O-DRL method that exploits $Q$ values, provided by function $f(\cdot)$ to rank importance of features (image pixels or objects). Interpretable models are realized by pixels' and objects' salience. The explanation function $\epsilon_l$ realizes surface representations of interpretations using visual means.

\subsubsection{Understanding agent actions using specific and relevant feature attribution}
\label{sec:4.4.5-ExplainYourMove}

Recognizing that perturbation-based approaches for RL,  as that proposed by S. Greydanus et al. in \cite{VisualizingAtariAgents} and  the O-DRL approach \cite{TransparencyandExplanationInDRL} in \ref{sec:4.4.4-TransparencyandExplanationInDRL}, tend to produce saliency maps that are not specific to the action of interest, N. Puri et al. in \cite{ExplainYourMove} propose SARFA to generate saliency maps that focus on explaining the specific action decided, balancing between specificity and relevance. 

SAFRA is a pertrubation-based approach for generating saliency maps focusing on capturing the impact of pertrubation only on the $Q$ value of the action to be explained ({\em specificity}), and downweighting features that alter the expected rewards of actions other than the action explained ({\em relevance}): Specificity is measured by capturing the relative change to the $Q$ value for the action to be explained w.r.t. other actions, and relevance by considering how the relative expected reward of taking some actions (other than the one explained) changes with a pertrubation. The harmonic mean of these two measures produces the {\em salience of state features}.

SAFRA in \cite{ExplainYourMove} has been evaluated qualitatively in  games (chess, Attari games and Go), showing that in comparison to O-DRL \cite{TransparencyandExplanationInDRL} and the approach  in \cite{VisualizingAtariAgents} produces more focused saliency maps. In addition, human studies on problem-solving chess puzzles show that saliency maps produced by SAFRA are less confusing than those produced by \cite{TransparencyandExplanationInDRL} and \cite{VisualizingAtariAgents}. SAFRA is compared to the  other approaches using a chess saliency dataset produced by human experts, showing superiority in identifying salient features and robustness in pertrubations.

Overall, SAFRA  \cite{ExplainYourMove} addresses the outcome explanation problem, focusing on visual tasks and aiming to identify state features whose salience is specific and relevant to the decided action. SAFRA exploits $Q$ values, provided by $f(\cdot)$ and interpretation is provided by the salience of state features. The explanation is surfaced using visual means, showing the salience of features. 
\\ 

Regarding the  power of salient maps to assess the degree to which hypotheses about features of the learned policy correspond to the semantics of RL settings, A. Atrey et al. in \cite{ExploratoryNotExplanatory} suggest that salience maps should be used as an exploratory rather than as an explanatory tool and propose a methodology grounded in counterfactual reasoning, focusing on issues of subjectivity, unfalsiability and cognitive biases on mapping salient features to semantic concepts and behaviours. In addition to that, saliency maps have been used also in combination with policy explanation methods in \cite{LocalAndGlobal}, investigating their role when put in the context of agents' strategies (discussed in Section \ref{sec:4.2.1-Highlights}).

\subsubsection{Autonomous Self Explanation of Interactive Reinforcement Learning}
\label{sec:4.4.6-AutonomousSelfExplanation}

Y. Fukuchi et al. in  \cite{AutonomousSelfExplanation}  propose Instruction-based Behavior
Explanation (IBE) to explain agent’s
future behavior within a sufficient temporal extent, while the policy is being learned in an interactive RL setting. In IBE, an agent exploits the
instructions given by a human expert (which may not indicate the actions to be performed) to explain its own behavior: This happens under the assumption that when the agent receives more rewards it is likely that it followed the instructions given. 

IBE consists of two steps:
(i) estimating the target of the agent’s actions by simulation: the target is the change in the environment state after a fixed temporal extent for explanations specified by agents' actions in $n$ steps,
and (ii) acquiring a mapping from the target to the instructions, through a k-means classifier, in order to explain the action target
based on the instruction signal given by a human expert. Authors report on experiments with humans in a game setting, using two  DQN agents, each at a different training stage.   The most important finding is that the environment changes must be chosen deliberately for assigning an explanation signal: The parameter $n$ that defines the number of agent's actions to a target state can not be fixed, while predictions in settings of high-complexity must be accurate. 

IBE \cite{AutonomousSelfExplanation} provides a local interpretation method, interpreting the decision of the agent in specific states, given a target state in a \textit{constant} temporal extent of $n$ time steps. The interpretation function is the IBE method itself, exploiting agents' target and instructions provided by experts. Thus, the RL method is treated as a dark box, where the function $f(\cdot)$ fetches instructions provided by experts, provided as representations of agent's behaviour. The classification of assessed targets to instructions provides a local interpretable model. The explanation logic is context and domain dependent: examples of visualisations are provided in \cite{AutonomousSelfExplanation}. As stated in \cite{AutonomousSelfExplanation}, prediction of changes is challenging in complex settings, and the temporal extent may need to vary in different contexts, even for the same agent.

\subsubsection{Explainable reinforcement learning through causal lens}
\label{sec:4.4.7-CausalLens}

P. Madumal et al. in  \cite{CausalLens} aim to generate explanations using causal chains from action
influence graphs, 
for model-free RL agents. %This work is the first on exploiting causal explanation models for generating contrastive explanations for RL and it is motivated by evidence that causal models  are shown to provide subjectively better explanations and better performance in task prediction than state-action based explanations.
This is accomplished by encoding causal relationships between variables of interest and learning a structural causal model \cite{StructuralCausalModel}  during RL. Action influence models
approximate the causal model of the environment
relative to actions taken by an agent. This approach
uses causal models to generate contrastive explanations
for \textit{why} and \textit{why not} questions,
%Given assumptions about the direction of causal relationships between variables, during the policy learning process this approach also learns structural equations, i.e., the quantitative influences that actions have on variables.
and formulates the minimally complete explanations, and minimally complete contrastive  types of explanations to be provided
from an action influence model.
%An explanation is defined as a pair that consist of:
%1) an explanandum, the event to be explained; and 2)
%an explanan, the subset of causes given as the explanation. 

A critical part of this approach is learning the structural equations, i.e., the quantitative influences
of actions on variables: Assuming that a DAG specifying causal direction between variables is given, structural equations  are approximated as multivariate regression models during the training phase of the RL agent, exploiting samples of agent interaction with the environment via an experience reply. 
Therefore, this is a mimicking process, where interpretable models  are structural equations specifying the effects of actions in  features, the causes of rewards, and the states resulting in rewards. These local  models $c_l$  provide local response and local objectives' explanations.   The explanation logic $\epsilon$,   provides explanations to the two types of questions: ``why" and ``why not".

This approach has been computationally evaluated on six RL
benchmarks domains using six different model-free RL algorithms with discrete actions.
Results indicate that the proposed models are robust and accurate
enough to perform task prediction with a negligible performance impact. A human study (120 participants) evaluated the method
in task prediction, explanation satisfaction,
and trust, showing that the proposed model performs better
than the tested baseline, but its impact on trust is
not statistically significant.

\subsubsection{Self-supervised discovery of causal features}
\label{sec:4.4.8-Self-SupervisedDiscoveryOfCausalFeatures}

S. Wenjie et al. in reference \cite{Self-SupervisedDiscoveryOfCausalFeatures} propose a self-supervised interpretable  framework (SSINet) to explain a trained policy model of vision-based RL agents to locate fine-grained causal features towards gathering state-specific and task-relevant  evidence for the agent's decisions.

Specifically, given a trained
policy model, the proposed framework learns an explanation model that 
predicts an attention mask. The  basic attention patterns learned are exploited for identifying the relative importance of features and  analysing failure cases. If the generated actions are
consistent when the policy takes as input either the state or the attention-overlaid state, the features highlighted
by the proposed method are considered to be task-relevant and
constitute evidence for the agent’s decisions.  The agent in \cite{Self-SupervisedDiscoveryOfCausalFeatures}  is pre-trained using model-free RL
algorithms including proximal policy optimization (PPO), soft actor-critic (SAC) and twin delayed deep deterministic policy
gradient (TD3).

More technically, the SSINet uses  a U-Net architecture, providing ``masked" states through a feature extractor, a mask decoder and a sigmoid non-linearity. SSINet uses the same feature extractor with the pretrained agent policy model, thus SSINet trains only the mask decoder. SSINet is trained using a dataset of state-action pairs generated by the policy.
Training is ``driven" by a mask loss function aiming to satisfy two desiderata: \textit{Maximum behaviour resemblance}, i.e.,  the agent's behaviour using masked states  must be as consistent as possible with that when using original states, and \textit{minimum region retaining}, i.e., attentions must attend as little information as possible.

SSINet has been evaluated on several Atari
2600 games, as well as on Duckietown, a
self-driving car simulator environment. Empirical results
verify the effectiveness of the proposed method, and demonstrate that
the SSINet produces high-resolution and sharp attention
masks to highlight task-relevant information. Special emphasis has been given to explaining why the agent performs well or badly quantitatively, by introducing evaluation metrics that assess the effectiveness of attention masks, in multiple RL algorithms and actor architectures.

In conclusion, the SSINet proposed in  \cite{Self-SupervisedDiscoveryOfCausalFeatures} is a  self-supervised machine learning process for vision-based DRL agents. Although it exploits samples from a  trained policy model, it does not aim to produce an interpretable model that mimicks or that distills the  knowledge acquired in the policy model: It rather  addresses the outcome explanation problem, providing a local model $c_l$ that identifies salient state features, which are further presented via appropriate visual masking techniques realized by $e_l$, applied in the original state.

\subsubsection{Distilling DRL in Soft Decision Trees}
\label{sec:4.4.9-DistillingDRLInSDTs}

Y. Coppens et al. in \cite{DistillingDRLInSDTs} illustrate
how Soft Decision Trees (SDT)  \cite{DistillingNNsInSDTs} can
be used as interpretable policy models. SDT are hybrid classification models of 
binary trees of predetermined depth, and  NNs. Each
branching node represents a hierarchical filter that
influences the classification of input data. The leaf nodes learn softmax distributions over possible classes using model parameters $\theta^{SDT}$, while  the overall loss is minimized using mini-batch gradient decent optimization. 

Using state action pairs generated from the  policy learned using the actor-critic Proximal Policy Optimization (PPO) algorithm, the soft decision tree is trained to classify states to actions. The applicability of the method
is evaluated on the Mario AI benchmark: Authors show  the fidelity with which SDTs can provide  interpretations, and elaborate on the performance
of these models when one would consider to use SDTs directly for task execution. 

This is a mimicking process (as authors in  reference \cite{DistillingDRLInSDTs} also state), despite the article title that characterizes the approach as a distillation process. The  model $c_l$ is the soft decision tree, which provides  local only interpretations, classifying states to agent actions. Despite the possibility of examining the learned filters along
any decision tree path from the root to leaf, it is questionable whether the soft decision tree offers a fully interpretable model, given the parameterized leaf nodes and the decision tree filtering nodes leading to an action distribution. As noted in the article, since the distributions in the leafs need to generalize and
provide an action strategy for multiple state samples, it is to
be expected that the resulting action distributions in the soft
decision trees can differ from the NN output of
the PPO agent for individual input samples: This results in low fidelity in mimicking the PPO policy. The interesting aspect of this work is that, bigger decision trees, offer greater agent rewards. However this comes with the cost of reduced interpretability. The explanation logic $\epsilon$ provides surface representations using hitmaps, exploiting the SDT model in spatial settings. The use of SDTs in non-spatial settings is something that worth further investigation.

\subsubsection{Memory-based explainable reinforcement learning}
\label{sec:4.4.10-MemoryBAsedXRL}

F. Cruz, R. Dazeley and P. Vamplew in \cite{MemoryBAsedXRL} propose
a memory-based explainable reinforcement learning (MXRL) approach according to which the agent  exploits an episodic memory to infer the probability of success and the number of transactions to
reach the goal state. Specifically,  the agent exploits past interactions with the environment, and by ``introspection" can compare the probability of choosing an action against the probability of being successful, thus providing explanations in terms of the necessity to complete an intended task.
The interpretable model aims to explain the agent's decision for an action at a specific state, based on past experience, exploiting transitions in successful runs. It is assumed that all transitions have been recorded. This limits the method applicability to small, discrete state - action spaces, such as the grid world in which experiments have been conducted.

In essence, this is a distillation process where the agent distills knowledge regarding the reasons that the RL agent favors specific actions for specific states, through past experience gathered in an ``interpretation" that applies at the episodic memory.  The explanation logic is provided via hitmaps and diagrams, showing probabilities for choosing actions in states, and probabilities to succeed in reaching a goal state.

\subsubsection{Enhancing explainability of DRL through selective LRP}
\label{sec:4.4.11-LRP}

As noted in \cite{Miller}, people usually prefer explanations that focus on selected, specific evidence, instead of showing every possible cause of a decision. Based on this insight, authors in \cite{LRP-argmax}  aim to adjust the layer-wise relevance propagation (LRP) \cite{LayerWiseRelPropagation} saliency map approach to focus on the parts of the input that are most relevant for the decision-making process of a DRL agent. Their contribution is twofold: (a) Their adjustment uses an {\em argmax} function to follow only the most contributing neurons, filtering out the most relevant information; (b) they adjust LRP to dueling DQN convolutional layers.

LRP describes a concept that can be applied to any classifier if that fulfills the following two requirements. First, it has to be decomposable into several layers of computation where each layer can be modeled as a vector of real-valued functions. Second, the first layer has to be the input $x$ of the classifier and the last layer has to be the real-valued prediction of the classifier $f(x)$. Any DRL agent fulfills those requirements, considering  actions decided as the output: In this case LRP can be used to address the outcome explanation problem. 
According to the LRP concept, relevance values $R^l_j$ are assigned to each computational unit $j$ of each layer 
$l$ in such a way that $R^l_j$ measures the local contribution of the unit $j$ to the prediction $f(x)$. The calculation of $R^l_j$ values follow the LRP concept, if it sets the relevance value of the output unit to be the prediction $f(x)$ and calculates all other relevance values by defining $R^l_j=\Sigma_{k \in \{j\textit{ is input to } k\}}R^{l,l+1}_{j,k}$ for messages $R^{l,l+1}_{j,k}$, such that $R^{l+1}_k = \Sigma_{j \in \{j\textit{ is input to }k\}}R^{l,l+1}_{j,k}$. Relevance values $R^l_j$ can be computed in a backward pass, starting from the output layer, towards the input layer. This concept has been applied to the dueling DQN architecture in \cite{LRP-argmax}, specifying the backward pass through the fully connected and the convolution layers. However to find the most relevant input neurons contributing to a specific decision, authors propose the {\em argmax} rule, defining the messages as follows:
\[
  R^{l,l+1}_{j,k} =
  \begin{cases}
    R^{l+1}_{k} & \text{if $j = argmax\{w_j a_j\}$} \\
    0 & \text{otherwise}
  \end{cases}
\]
where $w_i$ are model weights and $a_i$ the inputs.

This approach has been evaluated on three Atari 2600 games to verify that the saliency maps generated are more selective than the ones created by alternative LRP methods, while still including the information expected from visual explanations: This can be beneficial to the transparency of DRL methods.

Overall, this is an approach that addresses the outcome explanation problem, aiming to provide the most salient features towards specific DRL agents' decisions. The model constructed concerns the local contributions (relevance values) of inputs and neurons to the decision made, while  explanations are provided visually by means of inputs' saliency maps.

\section{Concluding Remarks}
\label{sec:Overall}

Explainability of closed-box methods that provide agents with the capacity to act autonomously in the real world, and particularly of  DRL methods, is rather challenging, albeit emerging: DRL, being successful to prescribe actions and courses of actions in complex settings according to the evolution of the environment, may comprise a number of closed-boxes, including models of the environment, of the agent objectives and  the agent policy, as well as models that fit states' or state-action pairs' values. Based on the general blueprint of DRL methods, in this article we formulate specific explanation problems: (a) The policy and objectives model explanation problem, aiming to provide interpretation of the overall - global - control logic behind specific agent's responses  prescribed by the agent's policy and motivated by the agent's objectives, in coarse granularity and large scales; (b) the responses and local objectives  explanation problem that aims to interpret agents'  timely - local - responses in specific states, also explaining the correlation between environment's state features with agents' objectives  in fine granularity and small scale;  and (c) the model inspection  problem, providing elements and properties of any DRL individual model. 

The article identifies specific paradigms that have been proposed for implementing XDRL methods, proposing a distinction between the distillation and mimicking approaches.  It proceeds to  provide a  review of state of the art methods for XDRL methods, addressing the needs of human operators - i.e., of those that take the actual and critical decisions in solving real-world problems. For each state of the art proposal the article describes the overall objectives and the specific explanation problem that is being addressed, the method used for providing interpretations, the interpretable models constructed according to the  DRL explainability paradigm followed, the explanation logic and the surface representations of explanations.

What can follow as a  general and straightforward remark  from this review, is that while there is a lot of work and progress towards solving XDRL problems, it is very early to  conclusions regarding DRL explainability and transparency: We need to understand the  possibilities to their full extent in all aspects across the pipeline, and in all different dimensions regarding the explainability desiderata. Indeed, there are many questions that remain unanswered, which present major challenges. Subsequent paragraphs try to point out the major and important ones, according to our view.

Starting from the questions that are stated in the introductory part of this article, there is not any work that identifies and provides evidence about the qualities of ``good" and objective explanations for RL agents. There is some evidence about the qualities of explanations from other scientific fields (e.g., for an overview in such evidence in social sciences one may start from \cite{Lipton}), but we need to bridge these to the effectiveness of RL agents' explanations (for any of the explanation problems stated), also considering the characteristics of the real-world contexts in which agents are deployed.  Specifically, while many approaches make specific proposals for interpretation and presentation of explanations, these are made in a fragmentary  and ad-hoc manner (e.g., exploiting few natural language templates, or hitmaps with specific characteristics) providing mixed-results or no evidence on the effectiveness of these explanations (maybe in combination with other modalities) in real-world settings. For vision-based RL, for instance, a feasible explanation approach is to learn t-Distributed Stochastic Neighbor Embedding (t-SNE) maps, while, there are works 
visualizing important features for RL agent’s
decisions using saliency maps: The role of these maps to the effectiveness of providing local explanations needs further investigation, also in combination with other methods that provide contextual and more broad perspectives of agent behaviour.
%============  ====================================================Only very few approaches evaluate explanations' in terms of their qualities, and as far as we know, no XDRL method is evaluated in terms of satisfying transparency requirements in a real-world setting. In addition to the above, there is not a single approach that provides a comprehensive XDRL framework that would support humans to examine  informative factors at different levels of detail, granularity and scale, especially in cases where a single explanation can not be given with one or two diagrams, or with a (potentially complex) instantiation of a natural language template. Finally, there is not any approach that proposes principles, guidelines, or a paradigm, on how explainability / transparency  can be integrated as an inherent feature of an agent from the early phases of its development. The generic XDRL framework described here and its instantiations based on the distinct identified paradigms for XDRL agents, provide a starting point for that. While few approaches follow the interpretable box design paradigm, aiming to design a system that is inherently explainable, most approaches propose solving a specific instance of the global/local model explanation problem exploiting state-action values, rewards received, and/or states' features. 

Interpretable machine learning presents many challenges (e.g. some of them are discussed in \cite{InterpretableML}), which are inherited to constituent models of deep reinforcement learning methods. However, RL agents that solve sequential problems need to provide explanations on sequences of decisions and exploit the constituent models' interpretations in intertwined ways. Bellow we present a list of  challenges that we  consider important to report essential progress in explainable deep reinforcement learning in particular: 

- \textbf{Build a toolbox for explaining DRL}: We need as general as possible and advanced tools for the surface representation of explanations from DRL agents: These must identify generic types of information to exploit and must include appropriate configuration functionality to provide information using multiple modalities, allowing users to explore explanations at different levels of detail, granularity and scale. Such a toolbox would  provide explanations' content exploiting specific types of information and transforming information (e.g., via projections, filtering, aggregation and extraction of features, etc.) in comprehensive ways, for any combination of problem dimensions.

- \textbf{Be as comprehensive as needed}: We need to realize approaches that interpret and explain the behaviour of DRL agents in as much as possible comprehensive ways: Such approaches should address combinations of the explanation problems stated, in a joined  way, supporting explaining the overall policy and objectives model, specific responses, indicating features importance in general, refining features' ranking in specific cases and inspecting the consequences, indicating parts of the state-action space of special interest at different levels of granularity and scale. This will allow synthesizing comprehensive explanations for humans to understand agents' decision making, tailored to specific context of agents' deployment, with respect to users' requirements, constraints and cognitive aspects.

-  \textbf{Bridge to theory, while answering pragmatic concerns}: We need to bridge properties/qualities of interpretable models and of explanations provided with fundamental theories and facts about human cognition and the qualities of explanations, so as to understand better  the mechanisms for providing context-dependent qualitative explanations.

- \textbf{Focus on transparency}: There is not any comprehensive XDRL framework which has been developed for providing explanations under pragmatic constraints. Although there are approaches targeting to transparency, the term is used more or less equivalently to interpretability. We need methods/frameworks whose explainability capabilities can be tuned to pragmatic constraints, focusing on  transparency.

- \textbf{Explore the interpretable box design paradigm}: We need to further explore the distinct XDRL  paradigms for interpretable box design, also in conjunction to other objectives (e.g. as in \cite{VIPER}): While the mimicking paradigm provides a solution that somehow by-passes explaining the deep models in a direct way, it adds a layer to the system that is susceptible to additional loss of accuracy in problem solving and raises questions about its fidelity to the control logic of the DRL agent. Models' distillation provides opportunities for generalizing effectively beyond the training examples provided originally to the agent and allow building  compact  models. However, we need to explore further  methods for distilling acquired knowledge in direct ways, and exploiting these models to provide concise explanations'  for different purposes and across problem tasks. 

- \textbf{Develop XDRL in a principled way}: We need principles, methodologies and tools for designing and developing XDRL agents from early development phases. While there are methods that follow the interpretable box design paradigm,  which is close to that aim, most approaches add explainability layers/modules to agents  with advanced DRL methods. 

- \textbf{Explore XDRL in complex and/or large-scale multi-agent settings}:  Solving explainability problems for multi-agent (MAS) and potentially complex settings, does not follow in a  straightforward way from solving these problems for individual agents: Interpretability may involve further filtering/aggregation steps in providing explanations' content, while we need to advance explainability and transparency in challenging but highly critical real-world MAS settings.

- \textbf{Set up experimental protocols with emphasis on pragmatic concerns}: We need setting up experimental protocols, methodologies and widely acceptable indicators regarding  how and what one should evaluate to provide evidence regarding humans' qualitative understanding and acceptability of explanations, closely connected to the functionality of DRL agents deployed in real-world settings. A.Holzinger at al. in \cite{ExplanationsQuality} propose System Causability Scale for measuring explanations' quality, that can be a starting point, incorporating issues of transparency.

- \textbf{Provide comparative evaluations for interpretability, explainability and transparency}: Although some kinds of models are considered more interpretable than others, or some explainability methods may provide better explanations than others  (whatever this means, but w.r.t the explanation problem addressed)  we need a comprehensive comparative evaluation on the quality of explanations that interpretable models can provide, also w.r.t pragmatic issues.

%\section{Evaluation of Interpretability methods for DRL}
%\label{sec:evaluation}

%\paragraph{Paragraph headings} Use paragraph headings as needed.
%\begin{equation}
%\end{equation}

%%%% For one-column wide figures use
%\begin{figure}
%  \includegraphics{example.eps}
%\end{figure}
%

\begin{acks}
This work has been partially supported by the TAPAS H2020-SESAR-2019-2 Project (GA number 892358) Towards an Automated and exPlainable Air traffic management (ATM) System. 
\end{acks}



%% References



%% New version of the num-names style
%\bibliographystyle{elsarticle-num-names}
%\bibliography{sample.bib}



\newpage
\appendix
\section{XDRL state of the art: Methods characteristics and evaluation}
\label{appendix}

This supplementary part describes in a succinct and comprehensive way the XDRL methods reviewed: The aim is to identify the main characteristics of each method, as these have been described in Section \ref{sec:stoA}, specifying the explainability problem each method aims to solve, the XDRL paradigm it follows, what is the explanation content provided, and how this is surfaced, according to the proposal made.
In conjunction to that information, and to further help identifying the most appropriate method to be applied in a specific context, this section provides XDRL requirements for  the DRL methods used, as well as ``XDRL - DRL interplay aspects": Access to DRL models' information and/or samples required, as well as whether interpretable XDRL models need to access this information in an online mode (i.e., while the DRL method is being trained, or while the DRL agent takes decisions) or in an offline mode (i.e., taking batches of required samples and data from  DRL models).

Specifically, three tables summarize the XDRL approaches: Table \ref{tab:overall} provides an overview of the approaches reviewed, starting from  methods addressing the model inspection problem, proceeding to those approaches that address the global policy and objectives' model explanation problems, and then, to those addressing the local response/objectives' explanation problem. Groups of methods are separated by a bold line. Table \ref{tab:overall} identifies per method the paradigm followed, the interpretable model constructed, the required access to the DRL models via the $f(\cdot)$ function  (i.e., the input to the interpretation process), as well as the output of the interpretation process - the explanation content, and finally the surface representation of explanations. 
Table \ref{tab:evaluation} provides an overview of evaluating each of the methods, focusing on  evaluation goals and assessing the  quality of explanations. It particularly specifies the involvement of humans in the evaluation process, while it provides information on the way explanations are evaluated and the objectives of that evaluation. It repeats the content of explanations provided per method from Table \ref{tab:overall}, in order to be made self-contained. Finally, Table \ref{tab:functionality} provides the profile of the DRL methods to which each XDRL approach is compatible with or with which it has been demonstrated/evaluated with. It particularly identifies the DRL method used in the article proposing the XDRL method (if any, else it identifies again the specific XDRL model); the DRL input assumed, as some of the approaches are assuming visual only input; the state and action space requirements; and finally,  whether XDRL is able to provide the explanation content in an online or in an offline manner, according to the distinction made above.

Important detailed remarks that can be made following the tabular presentation of methods, are as follows:

According to the mimicking/distillation distinction we made, the approaches that follow the distillation paradigm are underrepresented: Thus, we need more approaches that exploit DRL models (of whatever kind) in a direct way for providing explanation content and answering various questions in high fidelity, and at different levels of scale and granularity, addressing all explainability problems. 

Following this remark, attention models provide a promising approach towards this direction, but approaches using such models follow the interpretable box design paradigm, only. More importantly, attention models have been used  mostly  for visual agents (this holds for approaches that aim at solving the policy explanation and  the outcome explanation  problems). 

Abstractions of states, transitions and trajectories offer  promising generic approaches towards providing explanations at various levels of granularity and scale, and their applicability should be extended to arbitrary hierarchical levels of abstraction, exploiting also the possibilities of DRL methods to learn  hierarchical models.

Approaches that offer examples of agent's behaviour in appropriate contexts and under specific constraints, or that highlight important/critical aspects affecting agent's behaviour, have been proposed towards  building trust to agents' capabilities, or for training the humans on agents' capabilities, preferences and limitations. Such interesting approaches have been reported for the policy explanation, objective explanation and the outcome explanation  problems.

Finally, regarding evaluation of explanations, there are numerous objectives and ways to measure effectiveness that deem the approaches practically incomparable: This is a side affect of the lack of agreed definitions on terms and problems to be solved as far as explainability, interpretability and transparency are concerned.
Further on that, as far as  transparency is concerned, there is lack of an approach that aims at addressing pragmatic concerns to the deployment of XDRL agents, while the involvement of humans in experiments in any of the reported works does not concern the involvement of operators in (simulated) real world settings.

% Appendix \ref{appendix} provides further details on these tables.

%%%%%%%%%%%%%%%%%%%%%%%%%%%%%%%%%%%%%%%%%%%%%%%%
%%%%%%%%%%%%        TABLES
%%%%%%%%%%%%%%%%%%%%%%%%%%%%%%%%%%%%%%%%%%%%%%%%

\pagestyle{empty}
{\addtolength\textwidth{1cm}
\begin{landscape}

\setlength\LTleft{-4.6cm}

{\fontsize{6}{6} \selectfont 
\begin{longtable}{cScSSMS}

\caption{XDRL methods profiles}
\label{tab:overall}       % Give a unique label

\\ \hline \noalign{\smallskip}
\textbf{Method}  & \textbf{Problem addressed} & \textbf{Paradigm} & \textbf{Interpretable Model} & \textbf{Model aspects exploited}   \textbf{by }$f(\cdot)$ & \textbf{Explanation} & \textbf{Explanation}
\\

[Ref](Sect.) & & & ($c$ - model) & (arguments) & \textbf{content} &  \textbf{surface representation} \\
\noalign{\smallskip}\hline\specialrule{.2em}{.1em}{.1em}\noalign{\smallskip}

\cite{LIME}(\ref{sec:4.1.1-LIME}) & Model inspection (LIME) and  model explanation (SP-LIME) & Generic & linear model, decision trees, etc. &    Problem instances' features & (a) Important features (b) Examples of cases  & Visual\\ \hline

\cite{SHAP}(\ref{sec:4.1.2-SHAP}) & Model inspection and response explanation & Generic & linear model  & Problem instances' features & Important features  &   Visual\\ \specialrule{.2em}{.1em}{.1em}

\cite{Highlights}(\ref{sec:4.2.1-Highlights}) & Policy  explanation & Generic & Highlights (sub-trajectories with important states) & Trajectories, Q values & Sub-trajectories exemplifying agent behaviour   & Visual\\ \hline

\cite{LocalAndGlobal}(\ref{sec:4.2.1-Highlights}) & Policy  and outcome explanation & Generic & Highlights (policy) and LRP (Outcome) & Trajectories, Q values, NN weights & Sub-trajectories 
and saliency of features & Visual\\ \hline

\cite{ContrastiveExplanations}(\ref{sec:4.2.2-ContrastiveExplanations}) & Policy  explanation problem & Generic & Qualitative descriptions of policies rollouts & MDP, Q values, rewards & Trajectories in terms of user-interpretable descriptions of states and actions & Natural language (template-based)\\ \hline

\cite{CriticalStates}(\ref{sec:4.2.3-CriticalStates}) & Policy explanation & Generic & Critical states and corresponding actions & Q values & Critical states and actions & Visual\\ \hline

\cite{GrayingBlackBox}(\ref{sec:4.2.4-GrayingBlackBox}) & Policy explanation and inspection & Generic & Approximating the MDP via SAMDP  & State Features, rewards, neural activations of states & State  clusters' and SAMDP & Visual inspection of states' clusters and SAMDP via t-SNE\\ \hline

\cite{betterInterpretability}(\ref{sec:4.2.5-betterInterpretability}) & Policy explanation & Interpretable box design & Attention model  & None & State representation clusters' and reconstructed images for specific key-value pairs & Visual (state embeddings' clusters via t-SNE, and reconstructed images)\\ \hline

\cite{InterpretableOPE}(\ref{sec:4.2.6-InterpretableOPE}) & Policy explanation & Generic & Influence Functions & None & Influences of transitions  & Visual representation of trajectories and influences of transitions\\ \hline

\cite{AttAugmentedAgents}(\ref{sec:4.2.7-AttAugmentedAgents}) & Policy explanation & Interpretable box design & Attention model  & None & Input frames with attention maps & Visual (attention maps, saliency maps)\\ \hline

\cite{ConservativeQImprovement}(\ref{sec:4.2.8-ConservativeQImprovement}) &  Policy explanation & Interpretable box design &  Decision Tree & None & Decision Trees' constructs &   Not specified\\ \hline

\cite{HierarchicalinterpretableRL}(\ref{sec:4.2.9-HierarchicalinterpretableRL}) & Policy  explanation & Interpretable box design  & Hierarchical plans with  natural language descriptions of skills/tasks and Stochastic temporal grammar (STG) & None & Hierarchical plans with language-grounded tasks &  Natural language\\ \hline

\cite{LVIN}(\ref{sec:4.2.10-LVIN}) & Model explanation & Mimicking & High-level FSA and low-level MDP & Raw model (deep NN) weights & Transition matrices / graphs, and generated trajectories & Not specified\\ \hline

\cite{interpretableDRLwithLMUTs}(\ref{sec:4.2.11-interpretableDRLwithLMUTs}) & Model explanation & Mimicking & Linear Model U-Trees (LMUTs) & DRL decisions, $(s,a,r,s')$  and Q values & Decision making rules using partition cells and saliency maps & Not specified\\ \hline

\cite{PolicyLevelExplanations}(\ref{sec:4.2.12-PolicyLevelExplanations}) & Policy explanation & Mimicking & Abstracted Policy Graphs (Markov chains of abstract states) & Abstracted Policy Graphs and Features' importance & Conditional expected scores of states, Value function & Not specified\\ \hline

\cite{PIRL}(\ref{sec:4.2.13-PIRL}) & Policy explanation & Mimicking & Programmatic policies & None & Functional program & Functional program\\ \hline 

\cite{ExplainingByImitation}(\ref{sec:4.2.14-ExplainingByImitation}) & Policy explanation & Mimicking & Decision dynamics and decision boundaries & None & Action-belief trajectories & Visualization of trajectories in belief complexes\\ \hline

\cite{InterpretablePlanningStrategies}(\ref{sec:4.2.15-InterpretablePlanningStrategies}) & Policy explanation & Mimicking & Logical Program Policies (LPP) & None & Logical rules in disjunctive normal form  & Flawcharts representing the decision rules\\ 
\specialrule{.2em}{.1em}{.1em}

%(\ref{sec:4.3.1-ImprovingHuman-RobotH-RInteraction})\cite{ImprovingHuman-RobotH-RInteraction} & Objectives  explanation & Generic & Fitted human reward function & State features and latent variables & World and comprehension (latent) features  & Statements for manipulating comprehension features\\ \hline

\cite{CommunicateObjectives}(\ref{sec:4.3.1-CommunicateObjectives}) & Objectives explanation & Generic & Linear objective function parameters; Models of human beliefs on parameters & Examples of trajectories & Trajectories & Not specified\\ \hline

\cite{WhatDidYouThink}(\ref{sec:4.3.2-IntendedOutcome}) & Objectives explanation & Generic & Belief maps (discounted visitations of future state-action pairs) & Discounted $(s,a)$ visitations and Q-values & Belief maps  & Visualizations of belief maps (hit maps)\\ \hline

\cite{DistalExplanations}(\ref{sec:4.3.3-DistalExplanations}) & Policy and Objectives  explanation & Mimicking & Decision tree (policy) and causal model (objectives), opportunity chains (importance of responses) & None & Decision tree nodes, causal chain variables and distal action & Natural language\\ \specialrule{.2em}{.1em}{.1em}

\cite{RewardDecomposition}(\ref{sec:4.4.1-RewardDecomposition}) & Outcome explanation & Generic & Difference explanations & Q values & Decomposed Q values with reward difference and minimal sufficient explanations & Visual\\ \hline

\cite{AutonomousPolicyExplanation}(\ref{sec:4.4.2-AutonomousPolicyExplanation}) & Outcome  explanation & Generic & Control logic (states by means of variables and  actions by annotated functions & States' features & Representation of states' features & Natural language (template-based) using communicable predicates\\ \hline

\cite{TransparencyCommunicationforRL}(\ref{sec:4.4.3-TransparencyCommunicationforRL}) & Outcome explanation & Generic & POMDP components & Agent's beliefs, transition models, Q values & POMDP components (updates)  & Natural language (template-based)\\ \hline

\cite{TransparencyandExplanationInDRL}(\ref{sec:4.4.4-TransparencyandExplanationInDRL}) & Outcome explanation & Generic & Pixel/Object salience & Q values & Recognized objects' salience & Visual\\ \hline

\cite{ExplainYourMove}(\ref{sec:4.4.5-ExplainYourMove}) & Outcome explanation & Generic & Features' salience & Q values & Features salience (specificity, relevance) & Visual\\ \hline

\cite{AutonomousSelfExplanation}(\ref{sec:4.4.6-AutonomousSelfExplanation}) & Outcome  explanation & Generic & Target states and expert instructions & Experts' instructions & Mappings of changes to instructions & Visual\\ \hline

\cite{CausalLens}(\ref{sec:4.4.7-CausalLens}) & Response and Objectives  explanation & Mimicking & Structural equations (quantitative effects of actions to variables (approximated by multivariate regression models) & None & Causal chains in action influence models exploiting structural equations & Not specified\\ \hline

\cite{Self-SupervisedDiscoveryOfCausalFeatures}(\ref{sec:4.4.8-Self-SupervisedDiscoveryOfCausalFeatures}) & Outcome explanation & Mimicking & Attention masks highlighting important feature states & Policy-generated state-action pairs & Attention masks & Visual (applies to vision-based RL)\\ \hline

\cite{DistillingDRLInSDTs}(\ref{sec:4.4.9-DistillingDRLInSDTs}) & Outcome explanation & Mimicking & Soft Decision Tree & None & Learned filters along the traverse path in SDT  & Visual (hitmaps)\\ \hline

\cite{MemoryBAsedXRL}(\ref{sec:4.4.10-MemoryBAsedXRL}) & Outcome explanation & Distillation & Past transitions  & Q values & Probabilities of choosing actions and probabilities of success  & Visual\\ \hline

\cite{LRP-argmax}(\ref{sec:4.4.11-LRP}) & Outcome explanation & Generic & NN units relevance values & NN weights & Input relevance (saliency)  & Saliency maps\\ \specialrule{.2em}{.1em}{.1em}

\noalign{\smallskip}\hline

\end{longtable}
}

\restoregeometry

\end{landscape}}
\pagestyle{plain}

%%%%%%%%%%%%%%%%%%%%%%%%%%%%% EVALUATION %%%%%%%%%%%%%%%%%%%%%%%%%%%%%%%%%%%%%%%%%%%

\pagestyle{empty}
{\addtolength\textwidth{1cm}
\begin{landscape}

\setlength\LTleft{-3cm}

{\fontsize{6}{6} \selectfont 
\begin{longtable}{ccMMLM}

\caption{XDRL evaluation}
\label{tab:evaluation}       % Give a unique label
\\ \hline \noalign{\smallskip}
\textbf{Method}  & \textbf{Involvement} & \textbf{Content of explanations} & \textbf{Explanations' quality} & \textbf{Evaluation Objectives} \\

[Ref](Sec.) & \textbf{of humans} & & \textbf{measurement}& \\
\noalign{\smallskip}\hline\specialrule{.2em}{.1em}{.1em}\noalign{\smallskip}

\cite{LIME}(\ref{sec:4.1.1-LIME}) & Yes & (a) Important features (b) Examples of cases  &  Faithfulness and Trustworthiness given a set of gold features to be recovered & Increasing trustworthiness, improving classifiers\\ \hline

\cite{SHAP}(\ref{sec:4.1.2-SHAP}) & Yes & Important features  &  Consistency with human intuition, ability to explain subtle classification differences  & Sample efficiency; Understanding classifiers' abilities\\ \specialrule{.2em}{.1em}{.1em}

\cite{Highlights}(\ref{sec:4.2.1-Highlights}) & Yes & Sub-trajectories exemplifying agents' behaviour  & Helpful in assessing agents' capabilities and  trustworthiness & Increasing familiarity with agents' capabilities and limitations\\ \hline

\cite{LocalAndGlobal}(\ref{sec:4.2.1-Highlights}) & Yes & Sub-trajectories  and saliency maps & Modeling agents' behaviour, capabilities, attention and  trustworthiness & Building models of agents' behaviour, trusting agents and assessing the role of saliency maps in policy explanations\\ \hline

\cite{ContrastiveExplanations}(\ref{sec:4.2.2-ContrastiveExplanations}) & Yes & Trajectories in terms of user-interpretable descriptions of states and actions & Properties of explanations (length, amount of information, time-horizon of consequences) & Provide contrastive explanations in a comprehensive way\\ \hline

\cite{CriticalStates}(\ref{sec:4.2.3-CriticalStates}) & Yes & Critical states and actions & Subjective and objective development of humans' trust    & Establishing trust on agents' limitations and capabilities\\ \hline

\cite{GrayingBlackBox}(\ref{sec:4.2.4-GrayingBlackBox}) & No & State  clusters' and SAMDP visualizations & Effectiveness in analysing DQN's policy at different levels of abstraction & Analysing DQN's policy\\ \hline

\cite{betterInterpretability}(\ref{sec:4.2.5-betterInterpretability}) & No & State representation clusters' and reconstructed images for specific key-value pairs & Effectiveness in task execution comparable to deep Q-learning methods  & Interpretable visual Q-learning\\ \hline 

\cite{InterpretableOPE}(\ref{sec:4.2.6-InterpretableOPE}) & Yes & Influence of transitions &  Influence analysis  & Evaluating the trustworthiness of OPE methods for noisy and confounded observational data\\ \hline 

\cite{AttAugmentedAgents}(\ref{sec:4.2.7-AttAugmentedAgents}) & No  & Input frames with attention maps & Predictability, comprehensibility and focus of attention  & Interpretable visual DRL\\ \hline

\cite{ConservativeQImprovement}(\ref{sec:4.2.8-ConservativeQImprovement}) &  No  & Decision Trees' constructs &  Quality of the policy (modelled by a decision tree) an model size & Interpretable Q-learning \\ \hline

\cite{HierarchicalinterpretableRL}(\ref{sec:4.2.9-HierarchicalinterpretableRL}) & No & Hierarchical plans with language-grounded tasks & None & Policy learning, generalization and interpretability\\ \hline

\cite{LVIN}(\ref{sec:4.2.10-LVIN}) & No  &  Transition matrices / graphs, and generated trajectories & Not measured & Imitation; Interpretability, verifiability and manipulability of the model\\ \hline

\cite{interpretableDRLwithLMUTs}(\ref{sec:4.2.11-interpretableDRLwithLMUTs}) & No & Decision making rules using partition cells and saliency maps & Fidelity to the DQN model decisions, samping efficiency and effectiveness in performing tasks &  Interpretability and direct task performance \\ \hline

\cite{PolicyLevelExplanations}(\ref{sec:4.2.12-PolicyLevelExplanations}) & No & Abstracted Policy Graphs (Markov chains of abstract states) and Features' importance & Prediction of features' importance and explanation (model) size  & Concise policy-level explanations' model in tractable time \\ \hline

\cite{PIRL}(\ref{sec:4.2.13-PIRL}) & No & Functional program & Performant and short programmatic policies & Interpretability and verfication of programmatic policies \\ \hline

\cite{ExplainingByImitation}(\ref{sec:4.2.14-ExplainingByImitation})& Yes & Action-belief trajectories & Interpretability, Accuracy, Subjectivity & Understanding  decisions dynamics \\  \hline 

\cite{InterpretablePlanningStrategies}(\ref{sec:4.2.15-InterpretablePlanningStrategies}) & Yes & Logical Program Policies (Logical rules in disjunctive normal form) & Expected score of the plan and agreement withe conveyed policy  & Improving human planning quality and decision making by conveying policies as flowcharts\\  \specialrule{.2em}{.1em}{.1em}

%(\ref{sec:4.3.1-ImprovingHuman-RobotH-RInteraction})\cite{ImprovingHuman-RobotH-RInteraction} & Yes & State and latent features  & Subjective humans' satisfaction on justifications  & Helpfulness explaining failures and positive user experience\\ \hline

\cite{CommunicateObjectives}(\ref{sec:4.3.1-CommunicateObjectives}) & Yes & Examples of trajectories & None & Understanding agents' behaviour via modeling the objective function \\ \hline

\cite{WhatDidYouThink}(\ref{sec:4.3.2-IntendedOutcome}) & No & Belief maps (discounted valuations of future state-action pairs) & Not measured    & Consistency of intended outcome with Q values\\ \hline

\cite{DistalExplanations}(\ref{sec:4.3.3-DistalExplanations}) & Yes & Decision tree nodes, causal chain variables and distal action & Subjective evaluation of explanation quality and task prediction  & Understanding of the agent and provision of better explanations\\ \specialrule{.2em}{.1em}{.1em}

\cite{RewardDecomposition}(\ref{sec:4.4.1-RewardDecomposition}) & Yes & Decomposed Q values with reward difference and minimal sufficient explanations & None & Understanding agent decisions for delving into DRL issues \\ \hline

\cite{AutonomousPolicyExplanation}(\ref{sec:4.4.2-AutonomousPolicyExplanation}) & No & Representation of states' features & No measures: Subjective comparison with  experts' explanations & Interpretability of control logic \\ \hline

\cite{TransparencyCommunicationforRL}(\ref{sec:4.4.3-TransparencyCommunicationforRL}) & No & POMDP components & None & Interpretability and repairing trust\\ \hline

\cite{TransparencyandExplanationInDRL}(\ref{sec:4.4.4-TransparencyandExplanationInDRL}) & Yes & Recognized objects' salience & Correctness of matching and predictions agents' behaviour; consistency of explanations with saliency maps & Explaining the influence of objects in agents' decisions\\ \hline

\cite{ExplainYourMove}(\ref{sec:4.4.5-ExplainYourMove}) & Yes & State features' salience & Salience (qualitative), identifying specific and relevant features (accuracy), robustness to pertrubations & Understanding agent actions \\ \hline

\cite{AutonomousSelfExplanation}(\ref{sec:4.4.6-AutonomousSelfExplanation}) & Yes & Mappings of changes to instructions & None & Predictability of agents' behaviour \\ \hline

\cite{CausalLens}(\ref{sec:4.4.7-CausalLens}) & Yes & Causal chains in action influence models exploiting structural equations & Subjective evaluation for aggregate video explanations of explanation models & Better understanding of agent strategies and promoting trust \\ \hline

\cite{Self-SupervisedDiscoveryOfCausalFeatures}(\ref{sec:4.4.8-Self-SupervisedDiscoveryOfCausalFeatures}) & No & Attention masks & Feature Overlapping Rate and Background Elimination Rate & Enhanced explanation of visual agents' decision making process \\ \hline

\cite{DistillingDRLInSDTs}(\ref{sec:4.4.9-DistillingDRLInSDTs}) & No & Learned filters along the traverse path in SDT & None & Understanding the policy w.r.t to state features\\ \hline

%\textbf{Method}  & \textbf{Involvement of humans} & \textbf{Content of explanations} & \textbf{Explanations' quality measurement } & \textbf{Target task} \\ 

\cite{MemoryBAsedXRL}(\ref{sec:4.4.10-MemoryBAsedXRL}) & No & Probabilities of choosing actions and probabilities of success &  None & Explaining decision making w.r.t to the intended task\\ \hline

\cite{LRP-argmax}(\ref{sec:4.4.11-LRP}) 
& No & Relevance values of inputs &  None & Selective and inclusive saliency maps\\
\specialrule{.2em}{.1em}{.1em}

\noalign{\smallskip}\hline

\end{longtable}
}

\restoregeometry

\end{landscape}}
\pagestyle{plain}

%%%%%%%%%%%%%%%%%%%%%%%%%%%%% FUNCTIONALITY %%%%%%%%%%%%%%%%%%%%%%%%%%%%%%%%%%%%%%%%

\pagestyle{empty}
{\addtolength\textwidth{1cm}
\begin{landscape}

\setlength\LTleft{-2cm}

{\fontsize{6}{6} \selectfont 
\begin{longtable}{cMcSSS}

\caption{DRL methods characteristics}
\label{tab:functionality}       % Give a unique label
\\ \hline \noalign{\smallskip}

%\textbf{Method}  & \textbf{Task model} & \textbf{Task input}  & \textbf{State / Action Space} & \textbf{Learning Mode}  \\

\textbf{Method}  & \textbf{DRL model} & \textbf{DRL Input}  & \textbf{State / Action Space} & \textbf{XDRL Learning Mode w.r.t DRL}  \\

[Ref](Sec.) & & (Visual, etc)   & (Discrete / Continue) & (Online, Incremental or Offline) \\
\noalign{\smallskip}\hline\specialrule{.2em}{.1em}{.1em}\noalign{\smallskip}

\cite{LIME}(\ref{sec:4.1.1-LIME}) & Classification/prediction model & Generic &   Interpretable representations of state features &  Offline \\ \hline

\cite{SHAP}(\ref{sec:4.1.2-SHAP}) & Classification/prediction model & Generic &   Interpretable representations of state features &  Offline\\ \specialrule{.2em}{.1em}{.1em}

\cite{Highlights}(\ref{sec:4.2.1-Highlights}) & Policy  model & Generic & Discrete actions and states & Online\\ \hline

\cite{LocalAndGlobal}(\ref{sec:4.2.1-Highlights}) & Policy model & Visual & Discrete actions & Offline\\ \hline

\cite{ContrastiveExplanations}(\ref{sec:4.2.2-ContrastiveExplanations}) & Policy model & Generic  & Discrete actions & Online\\ \hline

\cite{CriticalStates}(\ref{sec:4.2.3-CriticalStates}) & No Model & Generic ( although visual seems more appropriate) & Discrete actions and state features & Online\\ \hline

\cite{GrayingBlackBox}(\ref{sec:4.2.4-GrayingBlackBox}) & Policy model in Q-Learning  & Generic & Discrete actions  &  Exploration \\ \hline

\cite{betterInterpretability}(\ref{sec:4.2.5-betterInterpretability}) & Policy model in Q-Learning (DQN) & Visual & Discrete actions  but discrete/continuous state features & Directed exploration \\ \hline

\cite{InterpretableOPE}(\ref{sec:4.2.6-InterpretableOPE})
& Policy model & Generic & Generic  & Offline \\ \hline

\cite{AttAugmentedAgents}(\ref{sec:4.2.7-AttAugmentedAgents}) & Policy model (LSTM) & Visual & Discrete actions  & Exploration\\ \hline

\cite{ConservativeQImprovement}(\ref{sec:4.2.8-ConservativeQImprovement}) &  Policy model (Decision tree) & Generic & Discrete actions and states &  Exploration \\ \hline

\cite{HierarchicalinterpretableRL}(\ref{sec:4.2.9-HierarchicalinterpretableRL}) & Hierarchical policy model   & Visual  & Discrete actions  & Curriculum learning\\ \hline

\cite{LVIN}(\ref{sec:4.2.10-LVIN}) & Policy model (CNN)  & Generic (restricted to 2D space) & Discrete actions & Offline\\ \hline

\cite{interpretableDRLwithLMUTs}(\ref{sec:4.2.11-interpretableDRLwithLMUTs}) & Deep policy model in Q-Learning (DQN) & Generic & Discrete actions & Online (active play) and Offline (batch mode) \\ \hline

\cite{PolicyLevelExplanations}(\ref{sec:4.2.12-PolicyLevelExplanations}) & Policy model & Generic  & Discrete actions and binary state features & Offline\\ \hline

\cite{PIRL}(\ref{sec:4.2.13-PIRL}) & Policy model (DDPG) & Generic & Discrete or continuous actions and states & Offline in periodic batches of histories via input augmentation\\ \hline

\cite{ExplainingByImitation}(\ref{sec:4.2.14-ExplainingByImitation})  & No Model & Generic & Generic & Offline\\ \hline

\cite{InterpretablePlanningStrategies}(\ref{sec:4.2.15-InterpretablePlanningStrategies}) & No Model & Generic & Generic & Offline\\ \specialrule{.2em}{.1em}{.1em}

%(\ref{sec:4.3.1-ImprovingHuman-RobotH-RInteraction})\cite{ImprovingHuman-RobotH-RInteraction} & Objectives  model & Generic & Generic (low dimensional) & Online \\ \hline

\cite{CommunicateObjectives}(\ref{sec:4.3.1-CommunicateObjectives}) & Linear objective function & Generic & Generic & Offline \\ \hline

\cite{WhatDidYouThink}(\ref{sec:4.3.2-IntendedOutcome}) & Value-based (Q-learning) & Generic (low-dimensional space with deterministic MDP) & Discrete state-action space & Online and Offline \\ \hline

\cite{DistalExplanations}(\ref{sec:4.3.3-DistalExplanations}) & Policy (Decision trees), distal action model and  objectives (causal model)   & Generic & Discrete Actions and states & Online\\ \specialrule{.2em}{.1em}{.1em}

\cite{RewardDecomposition}(\ref{sec:4.4.1-RewardDecomposition}) & Decomposed reward and policy (drDQN, drSARSA) & Generic & Discrete Actions & Online\\ \hline

\cite{AutonomousPolicyExplanation}(\ref{sec:4.4.2-AutonomousPolicyExplanation}) & Policy and domain model  & Generic & Discrete actions and discrete or continuous states & Offline\\ \hline

\cite{TransparencyCommunicationforRL}(\ref{sec:4.4.3-TransparencyCommunicationforRL}) & POMDP components (focusing on the Observations model) & Generic & Discrete actions and states & Offline\\ \hline

\cite{TransparencyandExplanationInDRL}(\ref{sec:4.4.4-TransparencyandExplanationInDRL}) & Policy model & Visual & Generic  & Online\\ \hline

\cite{ExplainYourMove}(\ref{sec:4.4.5-ExplainYourMove}) & Q values model & Visual & Discrete states and actions & Online \\ \hline

\cite{AutonomousSelfExplanation}(\ref{sec:4.4.6-AutonomousSelfExplanation}) & Policy model & Generic & Generic & Online \\ \hline

\cite{CausalLens}(\ref{sec:4.4.7-CausalLens}) & Policy model & Generic & Discrete actions  & Online \\ \hline

\cite{Self-SupervisedDiscoveryOfCausalFeatures}(\ref{sec:4.4.8-Self-SupervisedDiscoveryOfCausalFeatures}) & Policy model (PPO, SAC, TD3)& Visual & Generic & Offline \\ \hline

\cite{DistillingDRLInSDTs}(\ref{sec:4.4.9-DistillingDRLInSDTs}) & Policy model (PPO) & Spatial & Generic & Offline\\ \hline

\cite{MemoryBAsedXRL}(\ref{sec:4.4.10-MemoryBAsedXRL}) & Policy model (SARSA) & Generic & Discrete actions and states  & Offline \\ \hline

\cite{LRP-argmax}(\ref{sec:4.4.11-LRP}) & Policy model & Generic (applied to visual) & Discrete actions & Online \\ \specialrule{.2em}{.1em}{.1em}

\noalign{\smallskip}\hline

\end{longtable}
}

\restoregeometry

\end{landscape}}
\pagestyle{plain}
%%%%%%%%%%%%%%%%%%%%%%%%%%%%%%%%%%%%%%%%%%%%%%%%

\end{document}